\documentclass[fleqn]{article}
\usepackage{PRIMEarxiv}
\usepackage{bibspacing}
\usepackage{appendix}
\usepackage[utf8]{inputenc} 
\usepackage[T1]{fontenc}    
\usepackage[hidelinks]{hyperref}
\usepackage{url}            
\usepackage{booktabs}       
\usepackage{amsfonts}       
\usepackage{nicefrac}       
\usepackage{microtype}      
\usepackage{lipsum}
\usepackage{fancyhdr}       
\usepackage{graphicx} 
\usepackage{verbatim} 
\graphicspath{{media/}}     
\usepackage[notocbib]{apacite}
\usepackage{amsmath}
\usepackage{pgffor}
\usepackage{multirow}
\usepackage{float}
\usepackage{array}
\usepackage{mdsymbol}

\usepackage{MnSymbol}
\usepackage{rotating}
\usepackage{algpseudocode}
\usepackage{algorithm}
\usepackage{anyfontsize}
\usepackage{dsfont}
\usepackage{xcolor}

\usepackage{mathtools}

\usepackage{geometry}
\usepackage{rotating}
\newcommand{\ip}[2]{\parbox{#1}{\centering\vspace{0.25cm} #2}}

\newcommand{\cm}{$\checkmark$}

\title{{Neural Architecture Search for global multi-step Forecasting of Energy Production Time Series}}

\author{
  Georg Velev \\
  School of Business and Economics \\
  Humboldt University Berlin\\
  \texttt{velegeor@hu-berlin.de} \\
   \And
 Stefan Lessmann \\
  School of Business and Economics \\
  Humboldt University Berlin\\
  Bucharest University of \\
  Economic Studies\\
  \texttt{stefan.lessmann@hu-berlin.de} \\
}

\usepackage{acro}

\DeclareAcronym{EGF}{
  short = EGF,
  long = energy generation forecasting
}

\DeclareAcronym{LLM}{
  short = LLM,
  long = large language models
}

\DeclareAcronym{TLF}{
  short = TLF,
  long = total load forecasting
}

\DeclareAcronym{NAS}{
  short = NAS,
  long = neural architecture search
}

\DeclareAcronym{DAG}{
  short = DAG,
  long = directed acyclic graph
}

\DeclareAcronym{RL}{
  short = RL,
  long = reinforcement learning
}

\DeclareAcronym{MDP}{
  short = MDP,
  long = Markov Decision Process
}

\DeclareAcronym{AC}{
  short = AC,
  long = Actor-Critic
}

\DeclareAcronym{entsoe}{
short=ENTSO-E,
long=European Network of Transmission System Operators for Electricity
}

\DeclareAcronym{MCDM}{
  short = MCDM,
  long = multi-criteria decision-making
}

\DeclareAcronym{WV}{
  short = WV,
  long = walk-forward validation
}

\DeclareAcronym{GED}{
  short = GED,
  long = graph edit distance
}

\DeclareAcronym{OR}{
  short = OR,
  long = Operational Research
}

\DeclareAcronym{HPO}{
  short = HPO,
  long = hyperparameter optimization
}

\begin{document}

\setlength{\bibitemsep}{0.9\baselineskip plus 0.05\baselineskip minus .05\baselineskip}

\maketitle

\begin{abstract}
The dynamic energy sector requires both predictive accuracy and runtime efficiency for short-term forecasting of energy generation under operational constraints, where timely and precise predictions are crucial. The manual configuration of complex methods, which can generate accurate global multi-step predictions without suffering from a computational bottleneck, represents a procedure with significant time requirements and high risk for human-made errors. A further intricacy arises from the temporal dynamics present in energy-related data. Additionally, the generalization to unseen data is imperative for continuously deploying forecasting techniques over time. To overcome these challenges, in this research, we design a \ac{NAS}-based framework for the automated discovery of time series models that strike a balance between computational efficiency, predictive performance, and generalization power for the global, multi-step short-term forecasting of energy production time series. In particular, we introduce a search space consisting only of efficient components, which can capture distinctive patterns of energy time series. Furthermore, we formulate a novel objective function that accounts for performance generalization in temporal context and the maximal exploration of different regions of our high-dimensional search space. The results obtained on energy production time series show that an ensemble of lightweight architectures discovered with \ac{NAS} outperforms state-of-the-art techniques, such as Transformers, as well as pre-trained forecasting models, in terms of both efficiency and accuracy.  
\end{abstract}

\keywords{NAS, Time Series, Multi-step Forecasting, Energy Domain}

\section{Introduction}
The energy domain represents a dynamic field, which is currently characterized by notable changes towards renewable energy production, carbon footprint reduction using green energy sources, and efficient energy allocation facilitated by innovative grid technologies \shortcite{Trend_SmartGrid,Trend_RES_CarbonReduction}. The rapidly evolving nature of the energy domain makes accurate forecasting of energy generation both challenging and yet essential for maintaining the balance between supply and demand \shortcite{SupplyDemand_Balance}. Additionally, the deployment of lightweight prediction models ensures adherence to the operational requirements of resource-constrained energy systems, particularly in real-time or near-real-time applications. Overall, \ac{EGF} is concerned with predicting the amount of energy produced to meet the consumers' demand \shortcite{TLF_Def,EGF_Def}. The focus of \ac{EGF} is either a specific energy source, such as renewable energy, or the total amount of electrical energy produced at a particular period of time in a specific region.\\

Algorithmic advancements in the field of \ac{EGF} help minimize the mismatch between energy demand and supply. The boost in forecasting accuracy is particularly relevant for Demand Response Management techniques \shortcite{EF_Forecasting}. For instance, reliable energy generation predictions improve power grid stability and resilience as energy suppliers are better prepared for fluctuations in power demand. As a result, the increased flexibility to rapidly adapt to changing power circumstances facilitates the minimization of energy losses. Improvements in the energy resource allocation lead to less energy being used on the whole, which in turn reduces CO$_{2}$-emissions as well as transactional costs from electricity trading. For instance, accurate \ac{EGF} can optimize renewable energy usage by facilitating load shifting to time periods when renewable energy is available in excess.\\

The necessity for agile energy operating systems underlines the requirement for efficient time series forecasting models. We define efficiency as the total time required for model training and forecasting. The widespread adoption of renewable energy sources amplifies the relevance of close-to-real-time operational decisions \shortcite{RES_RealTime_Connection}. Specifically, the growing integration of green energy into power system results in an increase in the number of trades on the electricity spot market, especially in the last hour before the delivery starting point \shortcite{RES_ShortTermTrading,RES_ShortTermTrading_Report}. Thus, efficiency is essential for the application of \ac{EGF} models shortly before trade execution. Additionally, algorithmic efficiency is also relevant for dynamic tariffs. The latter mirror the current price on electricity exchange platforms, which is determined, among others, by the amount of renewable energy produced and available for trading \shortcite{DynamicTariffs_RES}. When there is a surplus of renewable energy, consumers benefit from lower electricity prices. Thus, algorithmic efficiency is vital for real-time predictions of renewable energy generation, as such estimates facilitate the instantaneous consumer load shifting.\\

While the field of \ac{EGF} necessitates both accurate and efficient time series methods, the current state-of-the-art models, i.e., transformers and pre-trained models for time series forecasting \shortcite{Benchmark_TS,Sundial}, primarily focus on prediction accuracy. In comparison to this, certain trend-and-seasonality decomposition-based approaches have proven more efficient than transformers, but they struggle to outperform recently published transformer-based frameworks \shortcite{PatchTST,Basisformer}. Concerning the model design, small architectural modifications could have a big impact on the outputs produced by deep learning methods \shortcite{PertubNAS}. Thus, the manual configuration of complex time series algorithms, which are capable of producing precise global multi-step forecasts efficiently, highly likely represents a very time-consuming and error-prone process. Moreover, prediction techniques tailored to energy time series must be continuously retrained and redeployed due to, e.g., fluctuations in power generation and dynamic grid topologies \shortcite{DynamicGrids}. Therefore, the performance generalization to unseen data is of particular importance in the energy sector. However, verifying performance robustness in a temporal context adds another layer of complexity to the task of designing novel time series forecasting methods. In this regard, \ac{NAS}, first introduced by \shortciteA{NAS_Beginning}, can mitigate these challenges because it enables the automated component-wise search of novel architectures tailored to a specific dataset type. Micro \ac{NAS} is concerned with discovering novel cell architectures, whereas macro \ac{NAS} utilizes existing cells to optimize the network architecture on the macro level. \shortciteA{NAS_Beginning} employ \ac{RL} during the search to sample micro and macro models from a high-dimensional search space. In the context of \ac{OR}, \ac{RL} methods are considered powerful alternatives to traditional dynamic programming techniques, which fail to discover optimal solutions for complex, multi-dimensional optimization problems \shortcite{DP}. Therefore, our research aims to develop a novel macro \ac{NAS} framework that achieves a balance between the forecasting accuracy, efficiency, and generalization power of the models discovered in an automated fashion using \ac{RL} for the global, multi-time-step \ac{EGF} in several short-term settings.\\

The first contribution of our research is the design of a novel macro \ac{NAS} search space that incorporates only efficient network components. These are also selected to capture the complex temporal patterns inherent in energy-related time series, e.g., local semantic information of the input sequences, multiple trend and periodic components, etc. Concerning the latter, our search space facilitates the sampling of both established periodic nonlinearities and novel activations, specifically tailored to the requirements of \ac{EGF}.\\

The second contribution of our study is the formulation of a novel reward signal, which our \ac{RL}-based search strategy optimizes while sampling macro \ac{NAS} networks. Our reward function accounts for challenges related to both performance generalization in the energy domain and the maximal exploration of different regions of the search space. The component associated with the former quantifies the scenario of overfitting to encourage the discovery of architectures that generalize well on out-of-time datasets. Additionally, the second term of our reward signal explicitly maximizes the diversity among the sampled architectures.\\

In an empirical context, the third contribution of our work is the discovery of an ensemble of \ac{NAS} models, which outperforms state-of-the-art transformer architectures as well as pre-trained models in terms of a multidimensional criterion that incorporates both computational efficiency and predictive accuracy. Additionally, our lightweight architectures, which \ac{NAS} discovers for the global multi-time step forecasting of the total amount of generated energy, achieve competitive performance when transferred to specific energy production types, e.g., renewable energy time series. This highlights the generalization power of our macro \ac{NAS} networks to unseen energy-related data.\\ 

The structure of our study is the following: Section \ref{sec:relatedWork} provides an overview of studies dealing with energy time series forecasting. Additionally, in this section, we also highlight limitations of existing \ac{NAS} frameworks for time series modeling. Section \ref{sec:MTH} presents details about the actor-critic framework for sampling novel macro architectures from the search space we define for \ac{NAS}. In Section \ref{sec:ER}, we present the results of our empirical research, and in the last section, we provide conclusive remarks.

\section{Related Work}
\label{sec:relatedWork}
This section presents an overview of studies dealing with energy forecasting as well as \ac{NAS} for time series prediction, as our research work lies at the intersection of these two fields. Thus, this summary establishes the motivation for incorporating specific time series models in our \ac{NAS} framework, which we describe in Section \ref{subsec:MMSS}, as well as for the selection of suitable baseline methods, which are detailed in Section \ref{subsec:NASTR}. Moreover, the overview of existing \ac{NAS} frameworks for global multi-step forecasting highlights research gaps in the automated design of complex time series models, as well as our contribution to the current state of the literature.

\subsection{Energy Forecasting}
\label{subsec:EF}
This section outlines different branches in the field of energy forecasting, emphasizing the prediction of energy generation. Additionally, we provide a comparative analysis of state-of-the-art time series forecasting techniques applied to energy-related data, highlighting their advantages and potential drawbacks in terms of efficiency and predictive accuracy. Furthermore, we underline the limited applicability of specific methods to various real-life scenarios due to shortcomings in the supported forecasting horizon settings.\\

Regarding the environmental impact of different energy sources, the literature on energy forecasting addresses the prediction of both non-renewable and renewable energy. The production of the former, e.g., fossil fuel energy, is known to leave non-reusable, radioactive residue, which in turn causes environmental pollution \shortcite{NonRE_FossilWaste,NonRE_NuclearWaste}. While this highlights the necessity to transition to renewable energy sources for power generation, \shortciteA{Natural_Gas_Survey} point out that approximately 20\%\ of the global energy demand is met through the production of fossil fuels. This has motivated the application of a diverse set of forecasting techniques for the modelling of non-renewable energy generation and consumption, e.g., statistical methods such as SARIMA as well as more complex nonlinear techniques such as SVM-based, MLP, and LSTM models \shortcite{NaturalGas_SARIMA,NaturalGas_SVM,NaturalGasConsumption_Review}.\\ 

In recent years, renewable power generation has gained significant attention in the energy sector due to its potential to minimize the negative impact, e.g., greenhouse gas emissions, of fossil fuel-based electricity production \shortcite{REReview_2023}. Concerning different renewable energy types, \shortciteA{RE_Classification} identify five main clusters: bioenergy, hydropower and geothermal energy, wind, and solar energy. \footnote{Bioenergy is generated through the thermal conversion of biomass to biodiesel, biogas, etc. Biomass represents a renewable, biological material extracted mainly from animals and plants. In comparison to this, the primary source of hydropower comes from the movement of water between higher and lower elevations. These movements enable the water to pass through a spinning turbine, resulting in hydroelectric energy generation. While geothermal energy is derived from the heat produced by decaying mineral sources inside Earth's interior, solar energy relies solely upon the heat and the light radiated by the sun.} The availability of solar and wind energy is strongly impacted by current weather conditions, e.g., varying wind speed, sunlight intensity, cloud coverage, etc. Thus, the variability in green energy sources makes them more unpredictable in comparison to fossil fuel energy \shortcite{RES_Variability_Unpred}. The challenges for integrating intermittent renewable energy sources into power grids have motivated a higher number of studies to explore the generation forecasting of wind and solar power than geothermal energy, hydropower, and biomass energy. The models applied for the generation estimation of renewable energy span from simple statistical methods relying on numerical equations with sustainability constraints to recurrent and convolutional neural networks, tree-based models, as well as transformer-based techniques \shortcite{BiomassForest_Statistical, BiomassLSTMCNN_Directly, ReservoirInflow_Malaysia, Solar_CNNRNNTr, Solar_RNNTr}. For more details on the literature in renewable energy production forecasting, we refer the reader to the overview provided by \shortciteA{RE_Survey} and \shortciteA{REReview_2023}.\\

Transformer networks, which are capable of modelling long-range dependencies in sequential data due to their distinctive (self-)attention mechanism, have been widely adopted in natural language processing, computer vision and time series modelling \shortcite{NLPTransformers_Survey,VisionTransformers_Survey,TimeTransformers_Survey}. They are commonly applied for the global, multi-step forecasting of energy-related data, e.g., \shortcite{PatchTST,Basisformer}. Time series transformers often implement algorithmic innovations to improve the predictive performance or the efficiency of the vanilla transformer introduced by \shortciteA{Vanilla_Tr}. For instance, both Autoformer and FEDFormer perform deep seasonal-trend decomposition using average pooling filters \shortcite{Autoformer,FEDformer}. Basisformer, which reportedly outperforms both Autofromer and FEDFormer on energy-related time series data, extracts latent trend and seasonal patterns that are consistent across the historical and the future view of the data using contrastive learning \shortcite{Basisformer}. In comparison to most time series transformers, Crossformer extracts both latent cross-time dependencies and cross-feature patterns \shortcite{Crossformer}. However, the two-stage attention blocks significantly increase the computational resources necessary to train the model. The current SOTA predictive performance on time series forecasting in multiple domains, e.g., in the energy domain, in the financial sector, etc., is achieved by PatchTST \shortcite{PatchTST,PatchTSTOcean,PatchTSTFinance}. Among the two main components of PatchTST, i.e., patching and vanilla multi-head attention mechanism, the former is responsible for the efficient extraction of partially overlapping subseries representations.\\

Concerning efficiency, MLP-based forecasting techniques, e.g., TSMixer, MTSMixer, etc., outperform transformers \shortcite{TSMixer,MTSMixer}. The main difference between TSMixer and MTSMixer lies in the factorization applied by the latter, which reduces the redundancy across both dimensions of the input time series. The ability to better filter out irrelevant temporal information has, in turn, a positive impact on the quality of the global multi-step forecasts. Additionally, a linear MLP-based decomposition approach, i.e, DLinear, offers significant advantages in terms of efficiency in comparison to transformers with decomposition blocks, e.g., Autoformer and FEDFormer \shortcite{DLinear}. Furthermore, \ac{LLM}-inspired zero-shot forecasting techniques eliminate the computational cost of the training process since these models are pre-trained on a large corpus of time series, and thus, can be directly applied for forecasting purposes \shortcite{TimesFM}. There are two main branches of \ac{LLM}-inspired models. The first branch, which consists mainly of probabilistic forecasting techniques, utilizes LLM components during the pre-training process, e.g., Chronos. A notable limitation of the latter is that this zero-shot forecaster is designed to estimate a maximum of 64 time steps in the future. This shortcoming makes the application of Chronos unreliable in settings that necessitate hourly forecasts for more than three days. By contrast, models such as GPT4TS, which utilize transformer decoders as their primary building blocks, can handle longer forecasting horizons, e.g., 96 and 192 time points \shortcite{GPT4TS}. The second branch of \ac{LLM}-inspired techniques represents pre-trained foundation models, which, unlike Chronos and GPT4TS, do not treat input time series as text. Several examples of recently released foundation models, which support forecasting horizons of varying lengths, include TimesFM, Sundial, and Time-MoE \shortcite{TimesFM, Sundial, TimeMoe}. A notable difference between these models lies in the fact that TimesFM and Time-MoE support deterministic forecasts, whereas Sundial produces probabilistic outputs. This is particularly relevant for efficiency, as deterministic approaches generally require less computational time for zero-shot forecasting purposes than distributional methods.\\

Overall, a noteworthy difference between pre-trained models and forecasting techniques, which require tuning of the trainable variables on a specific dataset, is that most of the former support univariate forecasts. In contrast, the latter facilitate global time series predictions generated efficiently within a single forward pass. Also, powerful pre-trained models often utilize millions of neural weights to produce forecasts, which significantly increases their capacity compared to methods that require dataset-specific training. Such differences motivate an in-depth performance comparison between pre-trained and traditional forecasting methods, considering both algorithmic efficiency and predictive error.

\subsection{NAS for Time Series Forecasting}
\label{subsec:NASTSF}
In this section, first, we present different machine learning tasks \ac{NAS} has been applied to in a time series context. Additionally, we provide a tabular overview of \ac{NAS}
studies dealing with global multi-time step forecasting of temporal data. Based on the overview, we identify research gaps in the field of time series forecasting and point out which \ac{NAS}-based frameworks are included as benchmarks in our empirical research. Also, we highlight the contribution of our study.\\

\ac{NAS}, initially introduced by \shortciteA{NAS_Beginning} for language modeling and image recognition, has since been applied to supervised and unsupervised time series tasks—including anomaly detection, e.g.,  \shortcite{NAS_Anomaly_1,NAS_Anomaly_2,NAS_Anomaly_3}, and spatial-temporal graph learning, e.g., \shortcite{NAS_graph_learning_1,NAS_graph_learning_2}—with growing interest in automated model design for classifying univariate and multivariate temporal data, e.g., \shortcite{NAS_Cl_1,NAS_Cl_2,NAS_Cl_3,NAS_Cl_4}. Our study focuses on \ac{NAS}-based discovery of lightweight models for the global multi-step forecasting of energy time series. Therefore, we exclude classification-specific \ac{NAS} frameworks due to differences in the modeling of the target variables, the choice of the loss functions, etc., while also deferring the evaluation of spatial-temporal graph components and their influence on the accuracy-efficiency trade-off in NAS-designed models to future research.\\

{\renewcommand{\arraystretch}{3.0}%
\begin{table}
\fontsize{13pt}{13pt}\selectfont
		\centering
		\resizebox{\textwidth}{!}{
			\begin{tabular}{|c|c|c|c|c|c|c|c|c|c|c|c|c|}
				\hline
				\multirow{5}{*}{\ip{3cm}{\textbf{NAS Method (Reference)}}}&  \multicolumn{11}{c|}{\textbf{NAS Methodology}} & \multirow{5}{*}{\ip{2.3cm}{\textbf{(Renewable) Energy\\ Generation Forecasting}}} \\
				\cline{2-12}
				& \multicolumn{9}{c|}{\textbf{Search Space}} & \multicolumn{2}{c|}{\multirow{2}{*}{\textbf{Objective Function}}}  & \\
				\cline{2-10}
				& \multicolumn{5}{c|}{\textbf{Layer Types}}  & \multicolumn{4}{c|}{\textbf{Macro-level Architecture}} & \multicolumn{2}{c|}{ }   &  \multicolumn{1}{c|}{ }\\
				\cline{2-12}
				&  \multicolumn{3}{c|}{\textbf{Efficient}} & \multicolumn{2}{c|}{\ip{3.2cm}{\textbf{Inefficient}}} &  \multirow{2}{*}{\ip{2.5cm}{\textbf{Efficient (Chain Structured)}}}  &  \multicolumn{3}{c|}{\ip{3.2cm}{\textbf{Inefficient}}} &  \multirow{2}{*}{\ip{3cm}{\textbf{Walk-Forward Validation}}} &  \multirow{2}{*}{\ip{3.5cm}{\textbf{Architecture Diversity Maximization (Graph Edit Distance)}}} & \\
			    \cline{2-6}\cline{8-10}
				& \textbf{MLP}  &  \textbf{CNN} & \textbf{Patching}  & \textbf{RNN}  & \ip{3cm}{\textbf{Multi-Head Attention}}   &   & \textbf{DAG}  &  \textbf{Encoder} & \textbf{Decoder}  &   &   & \\
				\hline
				\ip{3.5cm}{DRAGON \shortcite{DRAGON}} & \cm  & \cm  &  - & \cm  & \cm   & -  & \cm  & -  & -  &  - &  - & -  \\
				\hline
				\ip{3.5cm}{Online-NAS \shortcite{Online_NAS}} & -  &-   & -  & \cm  & -  & -  & \cm  & -  &  - & -  & -    &  \cm \\
				\hline
				\ip{3.5cm}{Hierarchical \\ \ac{NAS} \shortcite{Hierarchical_NAS}} & \cm  &  \cm & -  & \cm  & \cm  &  - & \cm  & \cm  &  \cm & -  & -   & -\\
				\hline
				Autopytroch-TS  &  \cm &  \cm & -  &   \cm&  \cm &  -  & -  & \cm  & \cm  & -  &  -  & - \\
				\hline
				AutoGluon-TS &  \cm & \cm  & -  &  \cm &   \cm & -  &  - &  \cm & \cm  & \cm  &-   & - \\
				\hline
				\ip{3.5cm}{DOCREL \shortcite{DOCREL}}&  - &  \cm &  - & \cm  &  -  & \cm  & -  & -  &  - &  - & -   & \cm \\
				\hline
				\ip{3.5cm}{SEANS \shortcite{SEANS}}& -  & \cm  & -  & \cm  & -   &  \cm &  - &-   & -  & \cm  & -   & \cm\\
				\hline
				\textbf{Our Study} &  \textbf{\cm} & \textbf{\cm}  &  \textbf{\cm} & -  & -   & \textbf{\cm}  &  - & -  & -  &  \textbf{\cm} & \textbf{\cm}  & \textbf{\cm} \\
				\hline
			\end{tabular}
		}
\caption{Overview of the methodology and the datasets used in \ac{NAS} studies dealing with global multi-time step forecasting.}
\label{tab:NAS_Overview}
\end{table}}

Table \ref{tab:NAS_Overview} provides an overview of \ac{NAS}-related studies for global multi-time-step forecasting. The columns related to \ac{NAS} methodology cluster the components included in the search space and the types of supported macro-level network architectures based on their efficiency. Additionally, Table \ref{tab:NAS_Overview} highlights in which cases the objective function of the studies accounts for performance generalization and maximal search space exploration. Moreover, for the sake of completeness, the dataset column facilitates the comparison of \ac{NAS} frameworks regarding their application for \ac{EGF}. The studies that consider renewable energy production data use only wind generation time series. To account for this limitation, our empirical research includes several renewable energy types, which are detailed in Section \ref{subsec:Data}. Concerning the search space components, Table \ref{tab:NAS_Overview} indicates that none of the studies rely solely on efficient layer types. The training and testing times are negatively impacted by RNN's recurrent computations, which require iterative looping through each sequence time step. While attention mechanisms do not involve recurrence operations, using multiple attention heads of the same type significantly increases the number of trainable variables that participate in the calculation of gradient updates and in the model testing step. In comparison to multi-head attention, patching splits the input sequences into subseries to extract local contextual information using efficient MLP-based computations. This prevents a drastic increase in the number of trainable variables in patching layers, while facilitating the extraction of new, latent features from observed subsequences. Concerning the impact of the macro-level architecture on algorithmic efficiency, Table \ref{tab:NAS_Overview} shows that only $\frac{1}{3}$ of the \ac{NAS} frameworks support the sampling of the well-known chain-structured models, the vertical depth of which grows with increasing number of layers. By contrast, in other macro-level architecture types the layers usually follow significantly more complex structure than the one produced by vertical stacking of network components. Consequently, the algorithmic complexity has a direct negative influence on the time necessary for model fitting and evaluation. Additionally, complex model design choices potentially result in overfitting. \ac{NAS}, which can produce very sophisticated architectures tailored to specific dataset requirements, is likely to discover models with low generalization power when the objective function of the search strategy lacks \ac{WV}-related components. Thus, evaluating the performance across multiple subsets is essential for selecting forecasting techniques that exhibit robust performance over time. Table \ref{tab:NAS_Overview} highlights the critical research gap, that most \ac{NAS} studies for time series forecasting do not incorporate \ac{WV} in their frameworks. Additionally, none of the approaches explicitly maximizes the architectural diversity of the sampled models. The automated discovery of robust architectures with high expressive power is a challenging task, which requires extensive exploration of different regions of the search space. Therefore, the maximization of the \ac{GED}, which measures the number of edits necessary to transform one architecture into another one, reduces the risk of getting stuck at local optimum points during \ac{NAS}. This, in turn, contributes to the quality of the discovered models.\\

Only a limited number of papers dealing with \ac{NAS} in domains different from time series forecasting describe potential ways of computing \ac{GED} between a pair of neural models. \shortciteA{NAS_EditDistance} define an optimal transport program that matches the layers in two \ac{DAG} networks to obtain a minimized value of the architectural distance. This involves assigning, e.g., convolutional layers with different kernel sizes to each other rather than to fully connected layers. While this approach works for search spaces where a single neural block consists of a single type of neural layer, it would not be the case for hybrid neural blocks containing, e.g., convolutional and fully connected layer types. Such hybrid model components facilitate efficient trend-seasonal decomposition, as detailed in Section \ref{subsec:MMSS}. Therefore, the so-called label penalty terms defined by \shortciteA{NAS_EditDistance} would be unsuitable for \ac{NAS} frameworks dealing with time series forecasting. Furthermore, quantifying the length differences between the shortest and the longest information flow paths between two \ac{DAG}-based networks would not be necessary for a search space that facilitates the sampling of chain-structured architectures. The flow of information in the latter does not necessitate distance-based computations due to its straightforward nature, as the output of each layer serves as the input to every subsequent layer. While the formulation of architectural similarity presented by \shortciteA{Autokeras} also relies on the idea of matching comparable layers, the authors do not normalize the distance to be in the range $[0,1]$. This has the disadvantage that the bigger the capacity of the sampled networks is, the higher the magnitude of the distance will be. This, in turn, could overshadow other components included in the objective function. The above presented specifics of existing \ac{GED} definitions make the reformulation of the distance between a pair of neural models essential in a time series context.\\

In the remainder of this section, we provide the motivation for including several of the \ac{NAS} frameworks presented in Table \ref{tab:NAS_Overview} in our set of baseline methods, among others. DRAGON, introduced by \shortciteA{DRAGON}, supports the modeling of computationally expensive \ac{DAG} neural models with a different number of nodes and varying connectivity. Different from chain-structured architectures, in \ac{DAG}s, deeper layers in the network can receive multiple inputs from preceding layers along the topological sort of the cycle-free graph. While incorporating four different layer types in DRAGON's evolutionary process ensures a flexible search space, the computational cost of the approach is also negatively impacted by recurrent and multi-head attention \ac{DAG} components. In contrast to DRAGON, Online-\ac{NAS} relies on recurrent computations only \shortcite{Online_NAS}. Hierarchical-\ac{NAS}, which represents gradient-based search, facilitates the modeling of an encoder-decoder architecture, in which each part of the macro-network can be \ac{DAG}-based \shortcite{Hierarchical_NAS}. Since hierarchical-\ac{NAS} incorporates the same layer types as DRAGON, hierarchical-\ac{NAS} is more computationally expensive than both DRAGON and Online-\ac{NAS}. Additionally, the objectives of these \ac{NAS} frameworks do not incorporate terms associated with \ac{WV} and the maximization of \ac{GED}. Therefore, our choice to include DRAGON, rather than Online-\ac{NAS} and hierarchical-\ac{NAS}, in the set of benchmarks for our empirical research is mainly motivated by DRAGON's advantages in terms of computational cost and the flexibility of the search space. Similar to hierarchical-\ac{NAS}, Auto-Pytorch-TS and AutoGluon-TS incorporate encoder and decoder macro networks in their search space. The main difference between these two frameworks and other \ac{NAS} methods is that they consider pre-defined macro architectures, e.g., PatchTST. The reason for including AutoGluon-TS in our list of \ac{NAS} benchmarks is three-fold. First, AutoGluon-TS supports the validation of the sampled models on multiple subsets, which is essential for the selection of robust forecasting techniques. Additionally, AutoGluon-TS supports several search modes, e.g., the efficient mode, which rules out deep neural architectures. Also, each search mode allows the specification of task-related constraints, e.g., exclusion of inefficient methods. Last, DRAGON's main competitor in terms of predictive performance is AutoGluon-TS, and not Auto-Pytorch-TS. The former outperforms the \ac{DAG}-based framework on 11 out of 27 time series datasets.\\ 

Analogously to our approach, two of the methods in Table \ref{tab:NAS_Overview} facilitate the automated discovery of chain-structured neural models \shortcite{SEANS,DOCREL}. However, we refrain from including SEANS\footnote{SEANS makes use of multiple objectives during the search process, one of which is devoted to maximizing the diversity of the trained weights. Both the sampled architectures and the initialization of each model impact the trained weights. Thus, maximizing the diversity among the already trained weight matrices does not necessarily guarantee the maximal exploration of different regions of the search space. This highlights the difference to our direct approach of minimizing the architectural similarity among the sampled \ac{NAS} models, which we detail in Section \ref{subsec:ACNAS}} and DOCREL in our benchmark set, as, neither of the methods offers an advantage over DRAGON's flexible search space.\\

Regarding the contribution of our study to the current state of the literature, Table \ref{tab:NAS_Overview} shows that our \ac{NAS} framework for global multi-step forecasting is the first to maximize the diversity among the sampled models, while also minimizing the risk of designing architectures that overfit a specific validation subset. Our formulation of \ac{GED}, which we detail in Section \ref{subsec:ACNAS}, builds on top of the existing work of \shortcite{NAS_EditDistance} and \shortcite{Autokeras}, while also overcoming the above-described limitations. Since discovering time series networks with high generalization power is daunting, our approach utilizes a flexible, high-dimensional search space, as described in Section \ref{subsec:MMSS}. The extensive exploration of different promising regions during the search also requires lightweight network building blocks. For this reason, we refrain from incorporating any inefficient components in the search process, as this would significantly increase the computational time of both the search and the deployment of the discovered models.

\section{Methodology}
\label{sec:MTH}
In this section, first, we present specifics about our high-dimensional search space for the automated design of global multi-step forecasting models. Afterward, we provide details about the sampling of entire macro networks using \ac{RL}. Additionally, we describe the role of the different components of our novel reward signal.
\subsection{Macro NAS Search Space}
\label{subsec:MMSS}
In this section, first, we provide details about the types of neural blocks we consider during \ac{NAS}. Also, we present a tabular overview of all macro-level network components included in our search space. \\

We incorporate five time series modules in the set of possible neural blocks to sample from for the automated design of chain-structured networks. Patching of temporal data facilitates the modeling of semantic local information by extracting subseries-level representations from a sequence of consecutive time steps \shortcite{PatchTST}. By capturing short-term dependencies in specific segments of the data, patching-based techniques can adapt to changes in local contexts and thus produce fine-grained forecasts of multiple time steps ahead. In addition to advantages related to predictive accuracy, a temporal patching module offers benefits in terms of algorithmic efficiency. For this reason, we design our patching module as follows:\\
\begin{equation} 
\label{eq:nonlinear_patching}
\begin{split}
\qquad\qquad\qquad\qquad
X_h & = \texttt{ReplicationPadding}(X) \\
X_h & = \texttt{Nonlinear\_Projection}(X_h)\\
X_h & = \texttt{Flatten\_Patch}(X_h)\\
X_h & = \texttt{Linear\_Projection}(X_h) \\
\end{split}
\end{equation}
The $\texttt{ReplicationPadding}$ layer, which is responsible for the segmentation into subseries-level patches, transforms the $3D$ input tensor with the shape $(batch\text{ }size, sequence\text{ }length,\text{ }channels)$ into a $4D$ tensor of the shape $(batch \text{ }size,$ $channels,\text{ }patches,\text{ }patch\text{ }length)$. The $\texttt{Nonlinear\_Projection}$ is applied along the last dimension of the $4D$ tensor to learn a higher-dimensional latent representation of the patches than their original length. Once the last two dimensions of $X_h$ are flattened, a linear projection layer maps the latent patch feature space to the length of the target window. Therefore, our patching block applies MLP-based transformations only along the temporal dimension of the input sequences. \\

Similarly, our second neural block, which performs an efficient nonlinear trend-seasonal decomposition, operates along the temporal dimension. The reason for including decomposition-based computations in our search space is to account for the complex patterns, which energy-related time series often exhibit, e.g., multiple trend and seasonal components, etc. \shortcite{Complex_Energy,Nonlinear_Trend_Energy}. Inspired by DLinear presented by \shortciteA{DLinear}, we model latent trend and seasonal components with nonlinear MLPs in the following way:\\
\begin{equation} 
\label{eq:nonlinear_trendseasonal}
\begin{split}
\qquad\qquad\qquad\qquad
X_{trend} & = \texttt{AvgPooling}(X) \\
X_{trend} & = \texttt{Nonlinear\_Projection}(X_{trend}) \\
X_{seasonal} & = X\minus X_{trend}\\
X_{seasonal} & = \texttt{Nonlinear\_Projection}(X_{seasonal})\\
X_h & = X_{seasonal} \plus X_{trend}\\
\end{split}
\end{equation}
The $\texttt{Nonlinear\_Projection}$ layers apply two potentially different activation functions, which we sample during \ac{NAS} for the modeling of each of the temporal components. We utilize average pooling with a specific kernel size to extract a linearly weighted representation of the trend. We also consider an alternative way of modeling the trend by replacing the first two computations in Equation \ref{eq:nonlinear_trendseasonal} with a nonlinear convolutional layer.\\

Given that the above-described network components do not model cross-dimension dependencies, i.e., the relationships among the input variables, we also include two variants of channel and temporal mixing neural blocks in our search space. MTSMixer accounts for the redundant information across both dimensions with factorized MLP computations \shortcite{MTSMixer}:\\
\begin{equation} 
\label{eq:mtsmixer_factorization}
\begin{split}
\qquad\qquad\qquad\qquad
X_{1},...,X_{f} & = \texttt{sample}(X) \\
X_{temporal} & = \texttt{merge(Nonlinear\_Projection}(X_{1},...,X_{f})) \\
X_{channel} & = X \plus X_{temporal}\\
X_{h} & = \texttt{Nonlinear\_Projection}(X_{channel})\\
\end{split}
\end{equation}
where $X_{1},...,X_{f}$ represent non-overlapping subsequences from $X$ selected  with tensor slicing operations. Each downsampled sequence undergoes a nonlinear MLP-based transformation before the merging operation restores the original temporal order. Since the nonlinear projection is applied to interleaved subsequences, rather than the entire original input sequence, the factorization across the temporal dimension aims to capture relationships among different time points with less redundancy. The nonlinear channel transformation reduces noise among the input variables by applying computations that bear some similarities to autoencoders. Specifically, the initial number of channels is first reduced to a lower dimension, which is then projected to the number of target time series. In addition to MTSMixer, we also include TSMixer introduced by \shortciteA{TSMixer}. The main difference between TSMixer and MTSMixer is that the former does not apply factorization.\\

Table \ref{tab:search_space_Overview} provides an overview of all components included in our search space. We set the maximal number of layers to three to avoid sampling very deep models, the training of which would substantially increase the running time necessary to complete the search. The dropout layers with the sampled rates are applied after the nonlinear projections in every time series block to prevent the macro networks from overfitting. Nonlinear projections make use of custom two-component activations only if the controller network samples an activation merging operation different from $none$. 
\begin{table}
\fontsize{6pt}{6pt}\selectfont
\centering
\begin{tabular}
{|c|c|c|}
\hline
\ip{2.0cm}{\textbf{Network Component}} & \textbf{List of possible Values} & \textbf{Conditioned on}\\
				\hline
				\ip{1.5cm}{Number of layers} & [2, 3] & - \\
				\hline
				\ip{2.5cm}{Time Series Neural Blocks} & \ip{3cm}{[Patching, Dnonlinear\_Avgpool,
				Dnonlinear\_Conv,
				MTSMixer, 
				TSMixer] }& - \\
				\hline
				\ip{1.5cm}{Dropout Rate} & [0.1, 0.2, $\ldots$, 0.5] & - \\
				\hline
				\ip{1.5cm}{Linear Prediction Projections} & [True, False] & - \\
				\hline
				Skip Connection & [1, 2] & \ip{3cm}{number current layer in [2, 3]}\\
				\hline
				\ip{3.0cm}{First and second Components of\\ first and second Activation Functions} & \ip{3.5cm}{ReLU, ELU, GELU, Mish, SiLU, Sine, Cosine, Tanh, Sigmoid, Linear, Exp, Erfcsoftplus} & \ip{2.5cm}{Second Activation sampled only if current neural
				Block in
                 }\\
                \cline{1-2}
                \ip{2.5cm}{Merging Operation for \\Components of both Activation Functions } &	[add, average, multiply, none] & \ip{2.5cm}{[Dnonlinear\_avgpool,
				Dnonlinear\_conv,
				MTSMixer, 
				TSMixer]}\\
				\hline
				\ip{2.5cm}{Number of hidden 
				Units} &	[200,300, $\ldots$,2000] & \ip{2cm}{current neural
				Block in 
				[Patching,
				MTSMixer, 
				TSMixer]}\\
				\hline
				Kernel Size & [3, 5, $\ldots$, 35] & \ip{2cm}{current neural
				Block in 
				[Dnonlinear\_avgpool,
				Dnonlinear\_conv]}\\
				\hline
				Stride & [4, 6, $\ldots$, 18] & \ip{2cm}{current neural
				Block = 
				Patching} \\
				\hline
                Patch Length & [8, 12, $\ldots$, 36] & \ip{2cm}{current neural
				Block = 
				Patching} \\
				\hline
			\end{tabular}
\caption{Overview of all macro network components included in our search space.}
\label{tab:search_space_Overview}
\end{table}
Patching is the only neural block among the five modules that does not utilize a second (custom) nonlinear function, as shown in Table \ref{tab:search_space_Overview}. Depending on the selected time series blocks, conceptually, the two (custom) activation functions can perform different types of transformations. While the first and the second activations are modeling latent trend and seasonal components produced by Dnonlinear, the two nonlinearities are responsible for the transformation of the temporal and the channel dimensions in the case of TSMixer and MTSMixer. Also,  Table \ref{tab:search_space_Overview} shows that in addition to well-known nonlinearities, e.g, $ReLU$, $GELU$, etc., we include $Cosine$ and $Sine$ activations in our search space to enhance the ability of the sampled networks to capture periodic patterns. Moreover, we include the recently introduced $Erfcsoftplus$ as an example for a hybrid rectifier sigmoidal nonlinearity \shortcite{Erfcsoftplus}.\\

Additionally, each neural block makes use of RevIN layers that \shortciteA{RevIN_Origin} introduce to combat distribution shift problems in time series forecasting. The RevIN modules standardize the input sequences, which are then fed to the sampled neural blocks, and denormalize the latent representation of the sequences for the final outputs. Since RevIN layers contain trainable parameters, they determine the amount of local statistical information to be removed from the nonlinear transformations of the inputs. The final predictions of the macro \ac{NAS} networks can either be produced by RevIN denormalization or by two MLP layers, which perform linear projections to the target channel and temporal dimensions.

\subsection{Actor-Critic Framework for Designing NAS Models}
\label{subsec:ACNAS}
In this section, first, we emphasize the role of \ac{RL} methods in \ac{OR}, and we clarify the connection between \ac{RL}-based \ac{NAS} and hyperparameter tuning. Afterward, we present specifics of the type of \ac{RL} algorithm, i.e., \ac{AC}, used for sampling macro-level architectures. Additionally, we provide details about the mathematical formulation of our novel reward signal, which incorporates an \ac{WV} term as well as \ac{GED}. Concerning the latter, we highlight differences between our definition of architecture similarity and the formulation presented by other \ac{NAS} studies. Also, we elaborate on the connection between the normalization of \ac{GED} and the exploration of various regions of the search space.\\

In \ac{OR}, a \ac{MDP} provides the mathematical framework for modeling sequential decision-making problems under uncertainty, optionally incorporating application-specific constraints \shortcite{MDP_Uncertainty,MDP_Constraints,MDP_Constraints2}. An \ac{MDP} characterizes a system that models the interactions between an agent and an environment. The latter produces rewards associated with new states as a result of the agent taking specific actions. These actions aim to maximize rewards collected by the agent in the long term. Dynamic Programming, which represents a traditional approach to solving \ac{MDP}s, is applicable as long as the problem scale remains manageable \shortcite{DP}. However, in scenarios characterized by high-dimensional problems within \ac{OR}, conventional methods fail to discover optimal solutions. For this reason, approximate dynamic programming techniques are utilized as an alternative approach to overcoming intractability challenges, which make finding exact solutions computationally infeasible. \ac{RL}-based frameworks, in which an agent, also called a controller network, interacts with an environment by sampling an action, are regarded as instances of approximate dynamic programming in the context of \ac{OR} \shortcite{RL_ADP,DP}. This is because deep \ac{RL} methods are capable of addressing high-dimensional sequential-decision making problems modeled as \ac{MDP}s.\\

In the context of \ac{NAS}, search strategies using \ac{RL} facilitate the sampling of novel architectures from a high-dimensional search space. Before we proceed to the specifics related to the search strategy employed in our empirical research, it is worth noting that \ac{RL}-based \ac{NAS} and classical \ac{HPO} paradigms can both be classified under the umbrella of automated machine learning \shortcite{AutoML}. 
\ac{HPO} refers to the automated configuration of hyperparameters for an arbitrary machine learning model, such as the number of trees in a random forest, the learning rate in gradient boosting methods, the batch size for fitting neural models, etc. \ac{NAS} extends \ac{HPO} by focusing solely on architectural design choices for configuring deep neural networks, in particular. While both \ac{NAS} and \ac{HPO} aim at discovering model configurations that minimize the residual with the target variable(s) the most, \ac{NAS} differs from \ac{HPO} in terms of the granularity and complexity of the search space defined for neural architecture components.\\ 

Our \ac{NAS} framework utilizes the \ac{AC} model to explore the high-dimensional search space defined in Section \ref{subsec:MMSS}. \ac{AC} represents an on-policy \ac{RL} algorithm, which incorporates a policy-based and a value-based component. The critic network produces the state-value function, which estimates the cumulative future reward that can be obtained from the current state of the environment. The actor network is associated with the action-selection policy, which facilitates the mapping of states to sampled actions. Regarding time series forecasting, the selected actions represent entire macro networks, whose components are tailored to handling temporal data. The estimation of the actor network’s gradients can be expressed with the following approximation formula:\\
\begin{equation}
\nabla_{\theta} J\big(\theta\big)\approx \sum_{t=1}^{T}\sum_{k=1}^{K}\nabla_{\theta }\text{log} P\big( a_{t,k}|a_{(t,k \minus 1)\text{:}1},s_{t\minus1}\text{;} \theta\big)\Big(\big( \sum_{t^\prime=t}^{T} \gamma^{t^\prime \minus t} R_{t^\prime}\big) \minus V^\pi_{t}(s_{t\minus1})\Big)
\label{eq:policy_gradient}
\end{equation}
where $\theta$ is related to the trainable parameters of the actor, and $K$ is the total number of network components that have to be sampled to build the macro-level architecture. $P$ is the probability for sampling each component conditioned on the previous state of the environment $s_{t\minus1}$, i.e., the macro architecture sampled in the preceding time step, and the previously selected network component within the same episode time step $t$. Additionally, the term $\big( \sum_{t^\prime=t}^{T} \gamma^{t^\prime \minus t} R_{t^\prime}\big)$ is associated with the so-called $Q$-function, which measures the sum of rewards to-go discounted with the factor $\gamma$. The difference between the $Q$-function and the state-value function produced by the critic, i.e., $ V^\pi_{t}(s_{t\minus1})$, is referred to as the advantage function. The latter highlights the connection between the actor and the critic networks. While the actor aims to maximize the probability of the most promising actions over time, the critic aims to minimize the difference between the estimated value function and the observed validation rewards, i.e., the advantage function. In other words, the critic guides the decisions of the actor during \ac{NAS} to discover model configurations that are most likely to forecast the values of all time series multiple steps ahead accurately. For more details on the role of the advantage function and the $Q$-function within the \ac{AC} framework, we refer the reader to Appendix \ref{sec:Appendix1}.\\

\begin{figure}[h]
\centering
\includegraphics[scale=0.70]{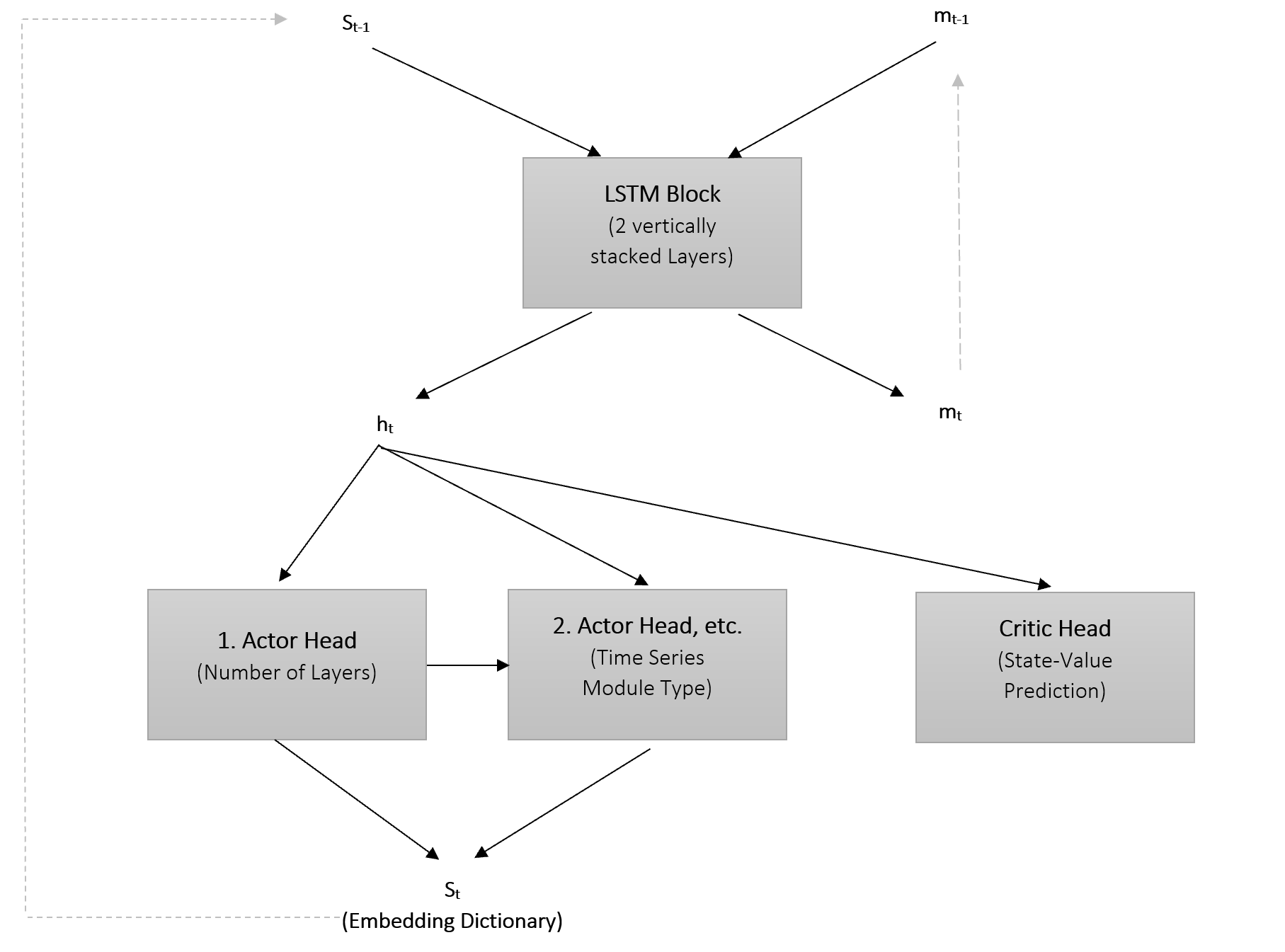}
\caption{Architecture of the agent and information flow within the actor-critic.}
\label{fig:AC_framework}
\end{figure}

In our study, the \ac{AC} implementation incorporates both the policy and the value component in the same neural network. Figure \ref{fig:AC_framework} visualizes the architecture of the controller network.
The two stacked LSTM layers represent the shared part of the network, which takes as inputs the state of the environment and the memory state produced by the recurrent block from the previous time step. While $m_{t\minus1}$ accumulates information about the sampled architectures within a single episode, we reset the memory state to a matrix of zeros at the beginning of each new episode. In this way, we prevent the controller from getting stuck at local optimum points due to accumulating unnecessary information from previous episodes. $S_{t\minus1}$ represents a continuous embedding matrix, the rows of which are associated with each component of the macro architecture selected in the previous episode timestep. Since the \ac{NAS} models are sampled from a conditional search space, as detailed in Section \ref{subsec:MMSS}, the state of the environment also includes continuous vectors indicating which of the conditional network components did not get sampled\footnote{We design a categorical search space, in which the sampling of specific components has conditional properties. Since neural networks cannot process categorical data, we initialize two embedding matrices. The first one contains the continuous representations of all non-conditional components, which get sampled for every macro network. The second matrix, which we name the "none"-embedding matrix, is meant for looking up the embeddings of conditional components, which could not be sampled in a specific episode time step. Embedding vectors from both matrices are used to produce sequences of unified length, which characterize the state of the environment at a given time point, i.e., $s_{t}$.}. In this way, from a technical perspective, the shape of the input sequences to the LSTM block remains constant throughout the entire search. Conceptually, we encourage the controller network to learn which \ac{NAS} components contribute to deteriorating forecasting performance, and should thus be avoided during the sampling process. The continuous vector representation of each categorical component is randomly initialized with 100 values drawn from a standard normal distribution. The LSTM block produces a nonlinear representation of the previously discovered network, i.e., $h_{t}$. The number of actor heads corresponds to the number of neural components that need to be sampled to create the entire macro network. Thus, sampling a single action in our \ac{NAS} framework amounts to sampling $K$ network components which are necessary to build the entire model architecture. While the first actor head takes only the environment's hidden state as input, the probabilities produced by every following prediction head are additionally conditioned on the embedding vector of the component selected by the preceding actor. In this way, the network architecture is generated in an auto-regressive way, as suggested by \shortciteA{NAS_Beginning}. The critic estimates the cumulative future reward based on the hidden state of the environment.\\

While the newly computed $S_{t}$ is used as the input to every following time step of the same episode, we adopt a slightly different strategy for the initialization of each new episode. We keep track of the best-performing model discovered during the search, and in 70\%\ of the cases we use its embedded representation at the beginning of an episode to encourage exploitation of promising regions. In the remaining 30\%\ of the cases, the episode start is conditioned on a randomly sampled macro network so that the controller explores new regions of the search space. For instance, in the initial stages of the search, the \ac{AC} network could leverage the complex seasonal components that energy time series exhibit by exploiting macro \ac{NAS} forecasting techniques with periodic nonlinearities. However, our search space facilitates the application of all activations in, e.g., both trend-seasonal decomposition blocks and time series layers specifically designed to filter out the noise from the input signal. Thus, exploration of new search space regions at the beginning of an episode would be essential to discover the configuration of the time series blocks that can best fit the complex temporal patterns of energy-related data. While our approach aims at achieving a balance between exploitation and exploration at the episode beginning, the architectures sampled towards the end of each episode are highly likely to share a lot of similarities, especially at later stages of the training process. Therefore, to increase the diversity of the sampled architectures during \ac{NAS}, we incorporate the \ac{GED} in the reward signal. We detail the novel formulation of the reward in the remainder of this section.\\

Equation \ref{eq:novel_reward} presents the three components of our novel reward signal:\\
\begin{equation}
\qquad\qquad\qquad\qquad
R_{t}=e^{\minus1}_{t,i}\plus wv_{t} \plus \frac{\sum^{t\minus1}_{t\prime = 1} ged(a_{t\prime}, a_{t})}{t\minus1}
\label{eq:novel_reward}
\end{equation}
where $e^{\minus1}_{t,i}$ is related to the inverse of the $RMSE$ computed between the true target values and the predictions on validation subset $i$, $wv_{t}$ is a penalty term quantifying overfitting in temporal context, and $ged(a_{t\prime},a_t)$ is the term associated with the architectural distance between the models sampled in the current timestep $t$, and the previous timestep $t\prime$. The \ac{WV} penalty term is further expressed as follows:\\  
\begin{equation}
\qquad\qquad\qquad\qquad
wv_{t}=\begin{cases*}
\minus 2 \cdot \big(e^{\minus1}_{t,i}\minus e^{\minus1}_{t,i\plus1}\big), \quad if \quad e^{\minus1}_{t,i}> e^{\minus1}_{t,i\plus1}\\
0, \quad otherwise
\end{cases*}
\label{eq:WV}
\end{equation}
where $e^{\minus1}_{t,i}$ and $e^{\minus1}_{t,i\plus1}$ are related to the rewards the \ac{NAS} models produce on the first and the second validation subsets $i$ and $i\plus 1$, respectively. When overfitting occurs, i.e., when the discovered models achieve a higher validation reward on the first subset than on the following one, we penalize the policy of the controller produced in the current time step by subtracting twice the difference in the rewards from Equation \ref{eq:novel_reward}.\\


The last term of our novel reward signal accounts for maximizing the architectural diversity among the sampled models. Depending on the initialization of the trainable variables of the controller network and the exploration of the environment at the beginning of the search, the agent may potentially become stuck in less promising local optimum points, unless explicitly incentivized to explore different regions of the search space by maximizing the \ac{GED} between discovered architectures. While our definition of \ac{GED} is inspired by the work presented in \shortcite{NAS_EditDistance} and \shortcite{Autokeras}, it also aims to overcome the limitations of existing formulations, as mentioned in Section \ref{subsec:NASTSF}. We eliminate the label penalty terms defined by \shortciteA{NAS_EditDistance} as they are not suitable for the computation of \ac{GED} between a pair of neural components using multiple layer types. Since our framework samples chain-structured \ac{NAS} networks, we also refrain from measuring the length differences between the shortest and the longest information flow paths between a pair of discovered models, as this is relevant only for \ac{DAG} networks. While we incorporate the non-assignment penalty terms from \shortcite{NAS_EditDistance} into our computation to account for unmatched layers, our normalization method, which expresses the edit distance in percentage terms, differs in that we use the minimum and maximum values of numerical network components as defined within our search space. Therefore, our approach depicts to what extent the agent selects \ac{NAS} models from maximally distant regions of the search space. Given two sampled macro architectures $a_{1}$ and $a_{2}$ with the layers $L_{1}\in a_{1}$ and $L_{2}\in a_{2}$, we formulate \ac{GED} as follows: \\
\begin{equation}
\qquad\qquad\qquad\qquad
ged(a_{1},a_{2})=\frac{ged_{m}(a_{1},a_{2}) \plus ged_{\char`\~m}(a_{1},a_{2})}{max(n_{1},n_{2})}
\label{eq:ged_formula}
\end{equation}
\begin{equation}
\qquad\qquad\qquad\qquad
ged_{m}(a_{1},a_{2})=\sum_{i\in L{1}, j \in L{2}}ed(l_{i},l_{j})
\label{eq:ed_matched_layers}
\end{equation}
\begin{equation}
\qquad\qquad\qquad\qquad
ed(l_{i},l_{j})=\frac{\sum_{z\in {l_{i}}, q \in l_{j}} d(c_{z},c_{q})}{n_{d}}
\label{eq:d_layers_components}
\end{equation}
where $ged_{m}(a_{1},a_{2})$ and $ged_{\char`\~m}(a_{1},a_{2})$ are related to the \ac{GED}s between all matched as well as all unassigned components, $n_1$ and $n_2$ are the number of layers in $a_1$ and $a_2$, and $ed(l_{i},l_{j})$ is the edit distance between a pair of matched layers. The term $ged_{\char`\~m}(a_{1},a_{2})$ assigns a distance of 1.0, i.e., 100\%\ , for every unmatched layer between two \ac{NAS} architectures. In the scenario of different layer numbers sampled for $a_{1}$ and $a_{2}$, one or several of the layers remain unmatched after the pairs with minimal edit distance are identified. Equation \ref{eq:d_layers_components} shows that the edit distance between a pair of layers involves the estimation of the average distance between the matching components $c_{z}$ and $c_{q}$ in the layers $l_{i}$ and $l_{j}$, respectively. Given that our search space has both numerical and categorical components, the distance between a pair of matching layer components can be expressed in the following way:\\
\begin{equation}
\qquad\qquad\qquad\qquad
d(c_{z},c_{q})=\begin{cases*}
\frac{| c_{z} \minus c_{q} |}{c_{max} \minus c_{min}}, \quad if \quad c \in \mathbb{Z}^{\plus}\\
1\minus \mathds{1}(c_{z} = c_{q}), \quad otherwise
\end{cases*}
\label{eq:components_distance}
\end{equation}
where $c_{max}$ and $c_{min}$ are the maximum and minimum values that are defined for the numerical components in our search space before performing the search. If one of the numerical components does not have an exact match, e.g., the patch length $c_{z}$ in the patching layer $l_{i}$ would not have a match in the trend-seasonal decomposition layer $l_{j}$, then $c_{min} = 0 $ and $c_{q}=0$. For categorical components, the percentual distance is expressed through a boolean comparison, which results in 100\%\ difference if two categorical components take on different values, and 0\%\ otherwise.
 
\section{Empirical Research}
\label{sec:ER}
\subsection{Data Retrieval and Data Preprocessing}
\label{subsec:Data}
In this section, we provide details about the data retrieval process of the energy production time series. Additionally, we describe the preprocessing steps we perform to transform the datasets into a suitable format for several forecasting settings using \ac{WV}.\\

We retrieved the two energy-related datasets for our empirical research for the period of time from 2023-09-05 until 2025-04-09 from the platform of the \ac{entsoe}. The latter provides free access to various energy-related datasets, including energy production time series. During \ac{NAS}, the sampled models are trained on the first dataset, which we refer to in the remainder of the paper as \ac{NAS} train dataset. Afterward, the best-performing models are also applied to the second energy production time series, i.e., \ac{NAS} transfer dataset, to explore the transferability of the discovered models. The \ac{NAS} train dataset contains the total amount of generated energy, i.e., the hourly sum of power generated by plants on both transmission and distribution system operators, for 19 European countries. The choice of which countries to include in the dataset was primarily motivated by the number of missing values in the time series available on \ac{entsoe}. The \ac{NAS} transfer dataset consists of several time series mostly related to renewable energy production, i.e., the generation of biomass energy, hydro energy within impoundment and run-of-river facilities, and onshore wind energy. Additionally, since approximately 20\%\ of the worldwide energy demand is still satisfied with non-renewable energy sources, as mentioned in Section \ref{sec:relatedWork}, we include a comparatively small portion of variables related to fossil fuel generation in our second dataset. The features in our second dataset come from three European countries, i.e., Spain, Germany, and Romania. The main reason for considering these countries is that they have onshore wind energy production facilities, which is due to their geological location within Europe. This facilitates the forecast of onshore wind energy generation for these specific countries.\\

\begin{figure}[h]
\centering
\includegraphics[scale=0.271]{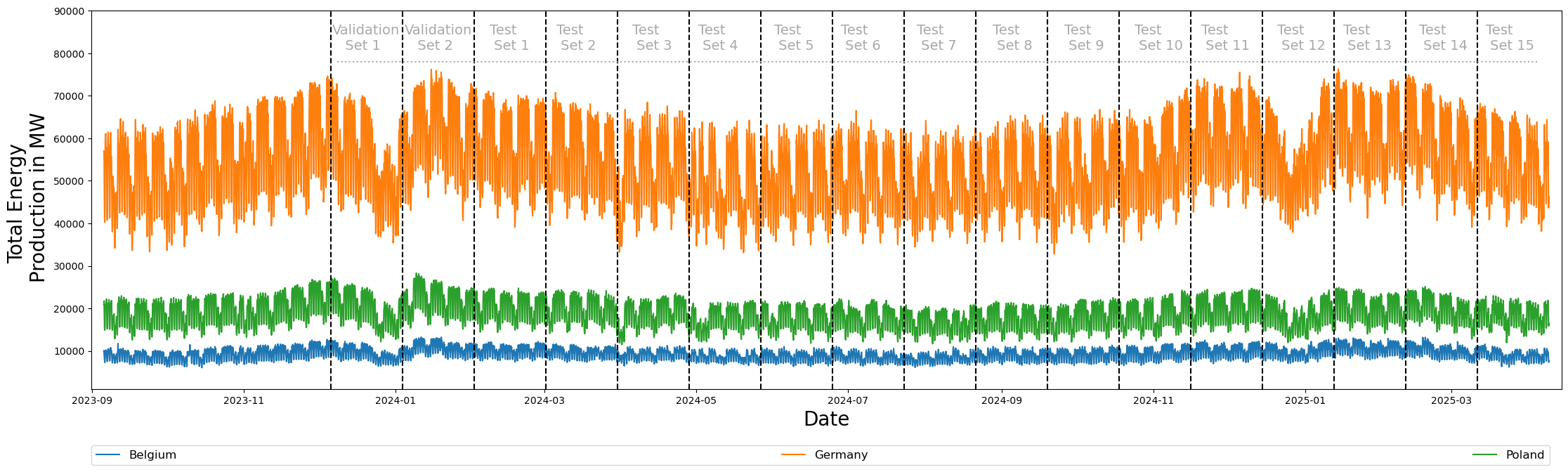}
\caption{Example for total energy production time series from Belgium, Germany, and Poland.}
\label{fig:TotalLoad}
\end{figure}

Figure \ref{fig:TotalLoad} visualizes the country-specific patterns of the total amount of generated energy per hour in two different regions of Europe, i.e., in Northwest and Central Europe. 
The dashed lines indicate the split into 17 subsets for walk-forward validation purposes. Due to the different magnitudes of the energy production per country, the data is standardized before the training process. The rescaling of the validation and test data is performed based on the mean and standard deviation statistics computed on the corresponding train subset before the time series datasets are split into sequences. We use the same number of time steps in the input and output sequences. In our empirical research, we focus on the short-term multi-step global \ac{EGF} in three prediction horizon settings $\{\ 48, 96, 192 \}\ $. The number of time steps in each train and test subset is chosen to yield a consistent number of sequences—1808 for training and 308 for testing—after splitting the time series into input and output windows for each prediction setting. The evaluation of the sampled \ac{NAS} models during the search uses the first two validation sets in Figure \ref{fig:TotalLoad}. Thus, the results in Section \ref{subsec:Res} are reported on the 15 out-of-time test sets.

\subsection{Training Details}
\label{subsec:NASTR}

We run \ac{NAS} for a maximum number of 1.000 episodes, each consisting of 10 time steps. In case the \ac{AC} network achieves average macro network probability per episode of 90\%\ or higher before the controller has sampled all 10.000 architectures, then the search is terminated. The macro \ac{NAS} models discovered within a single episode are trained and validated in parallel on the total energy production time series dataset for the three short-term forecast horizon settings $\{\ 48, 96, 192\}\ $. Thus, the error component from our reward signal in Equation \ref{eq:novel_reward} that we detailed in Section \ref{subsec:ACNAS} consists of the sum of $RMSE$ scores achieved on the validation subsets associated with the three prediction horizons. Since the residual component incorporates several error terms, we upscale the magnitude of the architecture edit distance 10 times to achieve a balance between the different components in our novel reward signal. The \ac{AC} network contains two LSTM hidden layers with 500 units each. The recurrent layers represent the shared part of the actor and the critic models. The learning rate of the controller is set to the low value of 0.0007 to prevent premature convergence. \\

We benchmark the predictive performance and the runtime of the discovered architectures against a total of 12 methods. First, to highlight the contribution of incorporating the \ac{GED} term and the \ac{WV} penalty in our novel reward signal, we provide the results from \ac{NAS} with a normalized entropy-based term. Similar to the \ac{GED} component of our novel reward signal, we upscale the entropy term 10 times. Overall, we replace the two components from our reward signal with the entropy term, as it aims to extend the exploration phase of the controller so that \ac{NAS} discovers macro networks with high expressive and generalization power. Since our search space includes time series blocks from the models TSMixer, MTSMixer, PatchTST, and DLinear, we also incorporate these four techniques in our baseline set. Additionally, we provide a comparison against Basisformer since the transformer model has been reported to outperform several recently introduced time series models, including DLinear \shortcite{Basisformer}. As mentioned in Section \ref{subsec:EF}, the \ac{LLM}-inspired zero-shot forecasting techniques GPT4TS, Time-MoE, TimesFM, and Sundial have been pre-trained on a large amount of time series data to forecast varying horizon lengths, including the prediction setting with the highest number of hours to forecast, considered in our empirical research, i.e, 192 hours. By contrast, Chronos is limited to predicting a maximum of 64 time steps. For this reason, among the five LLM-based and foundation models, we exclude only Chronos from our empirical research. We also provide a comparison against the \ac{NAS}-based technique DRAGON, as it represents a direct competitor of our approach w.r.t. to search space components, as mentioned in Section \ref{subsec:NASTSF}. Due to reasons related to computational resources, we limit the number of hidden layers for this \ac{NAS} framework to a maximum of eight graph nodes, which is still approximately three times higher than the maximum number of time series blocks to be sampled with our \ac{NAS} approach. Additionally, AutoGluon-TS represents the main competitor of DRAGON as pointed out by \shortciteA{DRAGON}. The \ac{NAS} method provides several modes, including a fast training setting that utilizes a search space of computationally efficient tree-based and statistical models. While we include AutoGluon-TS in its best-quality mode in our baseline models set, we exclude deep learning and tree-based models from the search space of AutoGluon-TS. In this way, we consider only efficient econometric methods during the baseline search, which still offer higher flexibility in terms of the hyperparameters to be tuned than the approaches included by default in the fast training mode of AutoGluon-TS. 

\subsection{Results}
\label{subsec:Res}
\subsubsection{\ac{NAS} Sampling History}
\label{subsubsec:SamplingHistory}
In this section, first, we provide episode-wise details about the search process that optimizes our novel reward signal as well as the entropy-based reward. Additionally, we present a two-dimensional representation of the regions of the search space visited by both controllers during \ac{NAS} to highlight general differences in the sampling history resulting from different formulations of the reward signal. Last, we shed light on the ramifications of the objective function design within \ac{OR} context.\\

\begin{figure}[h]
\centering
\includegraphics[scale=0.82]{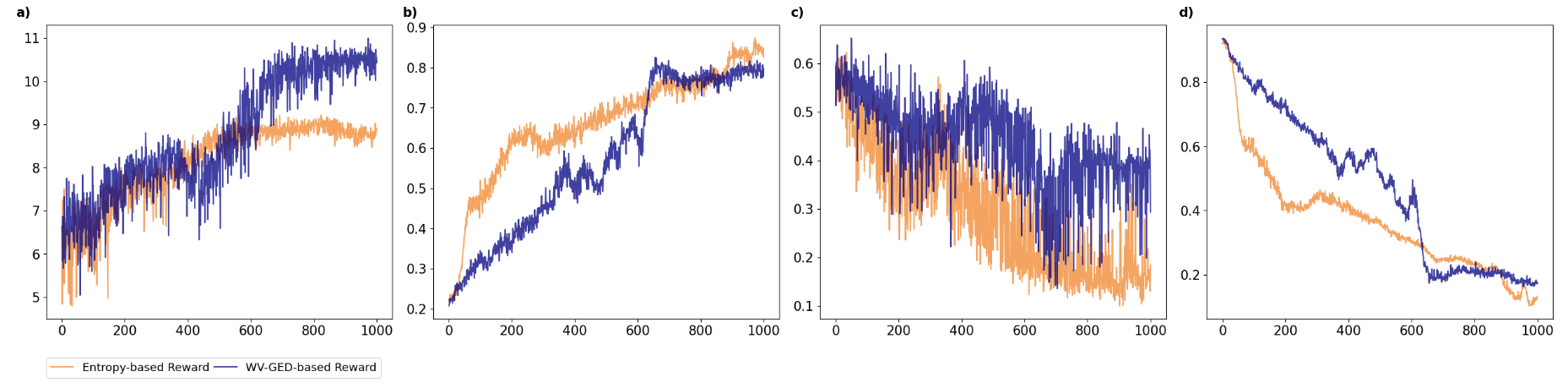}
\caption{\ac{NAS} training details related to the controller networks trained with our novel reward signal and the entropy-based reward, a) the rewards achieved per episode on average, i.e., the sum of the inverse of RMSE errors obtained on both validation subsets, b) the average \ac{NAS} model probability per episode, c) and d) the average \ac{GED} and the normalized entropy of the \ac{NAS} models per episode.}
\label{fig:NAS_Training_Details}
\end{figure}

Figure \ref{fig:NAS_Training_Details}a) shows that the rewards per episode differ mainly in the second half of the search. Between episodes 400 and 500, the controller trained with our novel reward signal, i.e., the WV-GED-based reward, visits regions of the search space associated with lower rewards, i.e., higher RMSE scores, compared to the entropy-based \ac{AC} network. We suspect that the reason for sampling \ac{NAS} models with lower predictive power during some periods of the search is related to the additional exploration of the search space encouraged by the WV-GED-based reward. This also involves exploring less promising search space regions. 
While the maximization of the architectural distance makes the WV-GED episode rewards noisier compared to the rewards achieved by the entropy-based controller, eventually, the WV-GED controller discovers search space regions associated with approximately 22\%\ higher rewards. The additional exploration also results in sampling more \ac{NAS} models with lower network probability on average than the architectures discovered by the entropy-based controller, as Figure \ref{fig:NAS_Training_Details} b) shows. Figures \ref{fig:NAS_Training_Details} c) and d) highlight that maximizing the architectural distance results in both higher \ac{GED} and, in most cases, in higher entropy in comparison to maximizing the uncertainty of the sampled models. We suspect the main reason for this counterintuitive finding is that during the search, we keep track of the embedding of the best-performing architecture that has achieved the lowest validation error so far, and use it to initiate the sampling of \ac{NAS} models at the beginning of each episode. While the \ac{AC} network is trained to maximize the entire reward signal as the controller reaches the final time steps of each episode, in the first episode time step, the \ac{NAS} models are sampled from regions characterized only by high predictive power. Put in different terms, we encourage the controller to maximize the WV-GED or the entropy term only if this would lead to improvements in the predictive performance of the sampled architectures. This is because our end goal is to find architectures with high expressive power, and not with high edit distance or entropy per se. Explicitly maximizing \ac{GED} leads to the continuous exploration of new potentially promising search space regions. By contrast, explicitly maximizing the sampling uncertainty does not directly impact how different the visited architectures would be, especially at the beginning of the search, when the actor heads produce low probabilities for most model components. This highlights the advantage of incorporating the architectural distance rather than the uncertainty component in the reward signal of the \ac{AC}. \\

\begin{figure}[h]
\centering
\includegraphics[scale=1.15]{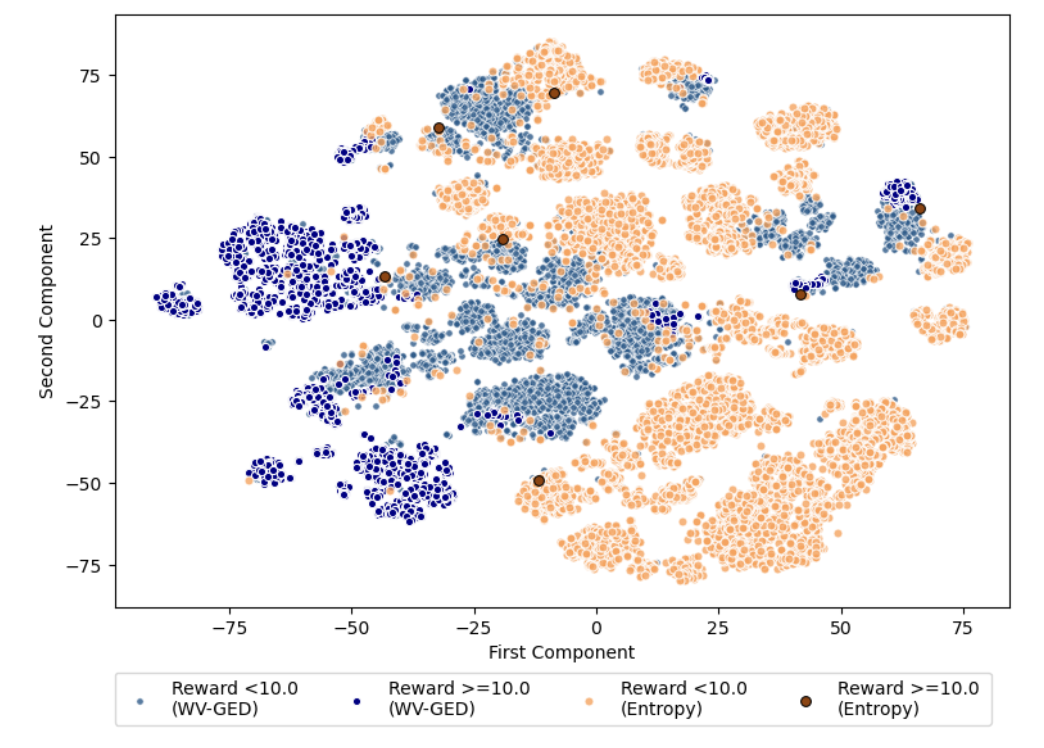}
\caption{Two-dimensional representation of the search space regions visited by both controllers during \ac{NAS}. The coloring of the scatterplot is related to the rewards, i.e., the inverse of the RMSE scores, achieved by the sampled architectures on both validation subsets.}
\label{fig:Explored_Search_Space_Regions}
\end{figure}

Figure \ref{fig:Explored_Search_Space_Regions} visualizes the two-dimensional t-SNE-based representation of the embeddings of \ac{NAS} architectures visited by both controllers.
The scatterplot shows that the \ac{WV}-\ac{GED} network indeed visits more distant regions of the search space during \ac{NAS} compared to the entropy-based controller, which samples most of the time series networks from the right search space region. The fact that the latter is associated with less promising architectures than the left search space region indicates that the entropy-based \ac{AC} has not managed to escape local optimum points discovered during the search. This is because the entropy reward signal does not explicitly incentivize the controller to maximize the architectural distance between the sampled models. This, in turn, results in sampling only seven models from a total of 10.000 visited architectures, which achieve a reward of 10 units or higher. In Appendix \ref{sec:Appendix2}, we show that some of the differences in the architectures discovered by both controllers are, e.g., related to the sampled time series blocks. While the most promising architectures visited by the WV-GED controller consist of MTSMixer blocks only, the best performing \ac{NAS} models discovered by the entropy-based \ac{AC} utilize convolutional-based trend-seasonal decomposition modules combined with MTSMixer blocks. While both controllers design composite nonlinearities with periodic activations, overall, the WV-GED \ac{AC} discovers more custom nonlinear functions. Generally, once each controller has sampled the first nonlinear component of a potentially composite nonlinearity, if no merging operation is selected in the following step, then the controller essentially selects an activation already defined in the search space. Since the entropy-based \ac{AC} is not explicitly encouraged to explore maximally distant regions of the search space, we suspect it is easier for it to sample well-established nonlinearities rather than to design novel composite activations. A further notable difference in the sampling history is that the best-performing \ac{NAS} models discovered with the WV-GED reward do not utilize linear prediction projection layers, which is the opposite of what is observed when maximizing the uncertainty in the visited architectures. \\

\begin{figure}[h]
\centering
\includegraphics[scale=0.97]{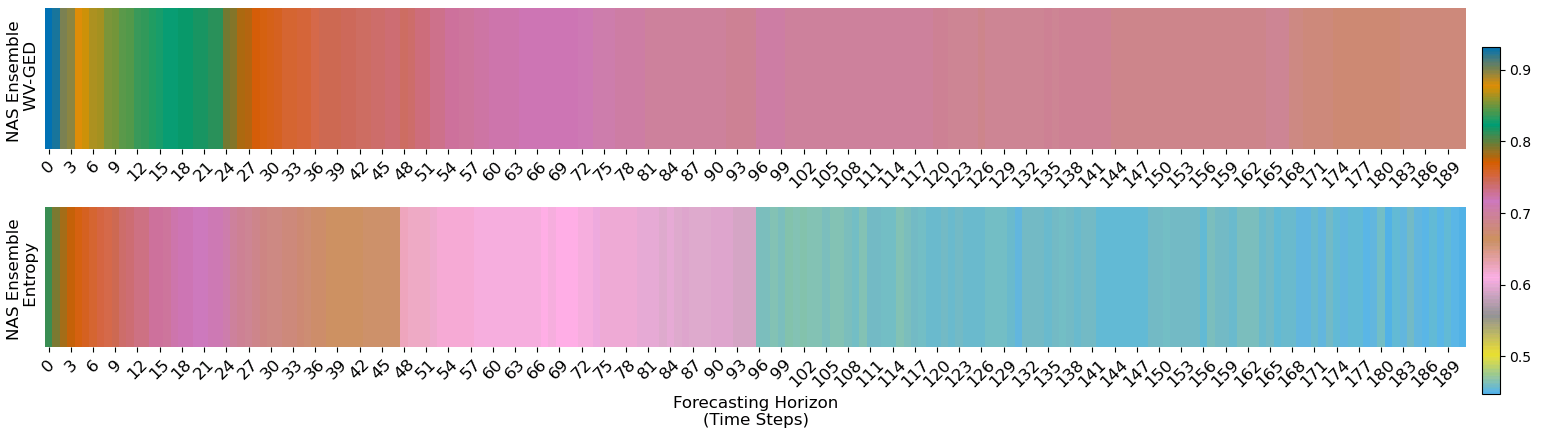}
\caption{Heatmaps visualizing the the Pearson correlation between the true values and the target predictions per time step obtained from the best performing \ac{WV}-\ac{GED}-based and entropy-based models. The correlation values are averaged across both time series datasets from  the three forecasting horizon settings $\{\ 48, 96, 192\}\ $.}
\label{fig:CorrelationTruesPredictions}
\end{figure}

The architectural differences in the macro time series networks visited during the search unavoidably impact the variation in the expressive power of the best-performing \ac{NAS} models discovered with both reward signals. The correlational heatmaps in Figure \ref{fig:CorrelationTruesPredictions}, which incorporate the average results from 10 runs for the three forecasting horizon settings $\{\ 48, 96, 192\}$, highlight the superior performance of \ac{WV}-\ac{GED}-based \ac{NAS} for \ac{EGF} in comparison to entropy-based \ac{NAS}.
Overall, the similarity between the actual values and the predictions obtained from entropy-based \ac{NAS} models drops approximately two times as the size of the prediction window increases. Consequently, for the time steps  $ [ 180, 181,... 192] $ the correlation falls to values close to 0.4, whereas the similarity between the actual values and the predictions produced by \ac{WV}-\ac{GED}-based \ac{NAS} remains approximately 0.20-0.25 units higher on average even in the most challenging cases.\\

The aforementioned differences, viewed through the lens of \ac{OR}, carry significant economic, operational, and ecological implications for applying \ac{NAS} to \ac{EGF}. Inaccurate energy forecasts introduce financial discrepancies in energy trading markets \shortcite{Economic_Impact_EGF_1,Economic_Impact_EGF_2,TSO_Balancing}. Limited exploration of maximally distant regions in the \ac{NAS} search space reduces forecast reliability, making the choice of reward signal critical for minimizing commitment errors in precomputed energy orders. Deviations between contracted and actual generation—particularly prevalent in renewable sources due to their intermittency—must be financially settled by energy generation entities. If \ac{NAS} models overestimate future output, transmission system operators would activate reserves via balancing markets and impose penalties on responsible actors. Consequently, traders employing entropy-based \ac{NAS} objectives face higher imbalance penalties than those using the proposed reward formulation. Conversely, underestimated \ac{NAS} forecasts may result in revenue losses in case surplus energy is sold at prices below the contracted rate.\\

Beyond market-related shortfalls, the choice of the reward signal also affects the reliance on costly backup generation systems. By shaping \ac{NAS} architectures, the formulation of the objective function directly induces predictive errors, which amplify discrepancies between predicted and actual generation, thereby heightening the need for sufficient flexible resources to reduce energy waste. Thus, entropy-based NAS models are expected to incur higher operational costs than WV-GED-based models, if power producers relying on the former decide to secure extra short-term storage or long-term contractual agreements with storage providers. The rising demand for energy storage capacity to buffer supply–demand fluctuations is especially critical for renewable sources \shortcite{CostlyEnergyBackup}.\\

Last, but not least, the design of the reward function in \ac{NAS} frameworks for \ac{EGF} bears substantial environmental ramifications. \shortciteA{CarbonBackup} highlight the unidirectional causal relationship from green to non-renewable energy sources resulting from the increase in installed fossil fuel capacity to meet unexpected demand, particularly during peak-load periods with low renewable output. Overall, the forecasting accuracy of energy production levels inherently deteriorates with increasing prediction window, as we showed in Figure \ref{fig:CorrelationTruesPredictions}. Thus, replacing the entropy-based term with the \ac{WV} and \ac{GED} components in the NAS objective can mitigate the dependence on carbon-intensive standby sources, especially in longer horizon prediction scenarios, by reducing the error in underestimated forecasts. 

\subsubsection{Predictive Performance and Efficiency of the discovered Architectures}
\label{subsubsec:PredictivePerfEfficiency}
In this section, first, we present the results in terms of predictive error from two different perspectives: the RMSE scores achieved on average per prediction horizon setting, as well as across the 15 out-of-time test sets. Additionally, we elaborate on the economic implications of different forecast deviations types produced by the most robust time series models. Afterward, we provide a ranking overview, which incorporates several error measures as well as a metric quantifying algorithmic efficiency, i.e., the training and testing time necessary to complete the evaluation of each time series model. Overall, the results reported in this Section are obtained from 10 model runs\footnote{We perform 10 runs with all models, which require dataset-specific training, to account for any variation coming from the initialization of the trainable weights. Additionally, we also perform 10 runs with those pre-trained time series methods, which produce distributional forecasts, to average out the sampling variation.}.\\

\begin{table}[h!]
\fontsize{23pt}{23pt}\selectfont
\centering
\renewcommand{\arraystretch}{2.1}
\setlength{\tabcolsep}{4pt}
\resizebox{\textwidth}{!}{
\begin{tabular}{|c|c|*{13}{c|}}
\hline
\multirow{2}{*}{\textbf{Dataset}} & \multirow{2}{*}{\textbf{\ip{4.0cm}{Prediction\\ Horizon}}} & \multicolumn{13}{c|}{\textbf{Time Series Model}} \\
\cline{3-15}
 & & \ip{2.5cm}{\textbf{NAS\\WV-GED}} & \ip{3.0cm}{\textbf{NAS\\Entropy}} & \ip{3.0cm}{\textbf{Dragon\\NAS}} & \textbf{AutoGluon-TS} & \textbf{DLinear} & \textbf{TSMixer} & \textbf{MTSMixer} & \textbf{Basisformer} & \textbf{PatchTST} & \textbf{Sundial} & \textbf{TimeMoe} & \textbf{TimesFM} & \textbf{GPT4TS}  \\
\hline
\multirow{3}{*}{\ip{2.5cm}{\ac{NAS} train\\dataset}} & 48h & \textcolor{orange}{0.446} & \textcolor{blue}{0.432} & 1.123& 0.802 & 0.691 & 0.545 & 0.837 &0.462 & 0.553 & 0.547& 0.544&0.608 & 0.995\\
\cline{2-15}
 & 96h & \textcolor{blue}{0.469} & \textcolor{orange}{0.473} & 1.139 & 0.711 & 0.650 & 0.565&0.862 & 0.514 & 0.557 & 0.525 & 0.522 &0.529 & 0.997 \\
 \cline{2-15}
 & 192h & \textcolor{blue}{0.506} & 9.313 & 1.261 & 0.668 & 0.637 & 0.621 & 0.937&0.592 & 0.542 & 0.556 & 0.547& \textcolor{orange}{0.521}& 0.990\\
\hline
\multirow{3}{*}{\ip{2.5cm}{\ac{NAS} transfer\\dataset}} & 48h & \textcolor{blue}{0.647} & 0.945 &  1.121 & 0.758& 0.756 & 0.658 &0.832 &\textcolor{orange}{0.656} & 0.664& 0.695 &0.670 &0.722 & 0.859\\
\cline{2-15}
 & 96h & \textcolor{orange}{0.708} & \textcolor{blue}{0.658} & 1.111 &0.806 & 0.792 & 0.705 &0.876 &0.720& 0.727 &0.745 &0.756 & 0.763& 0.918\\
 \cline{2-15}
 & 192h & \textcolor{blue}{0.786} & 0.885 &  1.356 &0.863 & 0.905 & \textcolor{orange}{0.806} & 1.007&0.822 &0.816 & 0.831 &0.867 &0.864 & 0.990\\
\hline
\end{tabular}
}
\caption{Results on both time series datasets for each prediction horizon setting. The numbers marked in blue are related to the lowest RMSE score, and the values marked in orange are associated with the second-best performing model.}
\label{tab:results_prediction_horizon}
\end{table}
Table \ref{tab:results_prediction_horizon} shows that most models achieve lower RMSE scores on the prediction of total supplier energy generation, i.e., NAS train dataset. We attribute this to the fact that most time series in the \ac{NAS} transfer dataset are related to renewable energy sources, the predictability of some of which, e.g., wind energy, is negatively affected by their intermittent nature. Additionally, the models discovered with our \ac{NAS} framework using the WV-GED reward signal represent either the first or the second best performing methods in comparison to all benchmarks across all settings. This highlights the methodological contribution of our research work to the field of global time series forecasting. While \ac{NAS} using WV-GED reward does not always outperform \ac{NAS} with an uncertainty-based reward signal, the latter produces more than 10 times higher error than most models included in Table \ref{tab:results_prediction_horizon} on the prediction of 192 hours of total energy generation. Overall, the maximization of the architectural distance results in sampling most of the well-performing time series models in later stages of the search. By contrast, some of the best-performing techniques discovered through the optimization of the entropy-based reward were visited by the controller in the initial stages of the training process. During the latter, the actor probabilities for most components are uniformly distributed. Thus, visiting such architectures amounts to randomly sampling time series models. For this reason, we suspect that by random chance, entropy-based \ac{NAS} discovers an ensemble of architectures which in some of the less challenging settings can fit the data slightly better than the methods designed with WV-GED \ac{NAS}. However, the results in Table \ref{tab:results_prediction_horizon} clearly show the opposite tendency as the prediction horizon increases to 192 hours. Regarding the \ac{NAS} benchmarks of our approach, we attribute the superior performance of AutoGluon-TS over Dragon \ac{NAS} to the incorporation of \ac{WV} in the search process. The two \ac{NAS} benchmarks are validated on the same time span. The difference with AutoGluon-TS lies mainly in that the approach splits the validation data into multiple walk-forward validation windows, which facilitate the selection of models that are less likely to overfit a single subset of data points.
\begin{table}[h!]
\fontsize{22pt}{22pt}\selectfont
\centering
\renewcommand{\arraystretch}{2.1}
\setlength{\tabcolsep}{4pt}
\resizebox{\textwidth}{!}{
\begin{tabular}{|c|*{13}{c|}}
\hline
 \multirow{2}{*}{\textbf{\ip{2.5cm}{Out-of-Time\\ Test\\Subset}}} & \multicolumn{13}{c|}{\textbf{Time Series Model}} \\
\cline{2-14}
& \ip{1.5cm}{\textbf{NAS\\WV-GED}} & \ip{2.7cm}{\textbf{NAS\\Entropy}} & \ip{2.3cm}{\textbf{Dragon\\NAS}} & \textbf{AutoGluon-TS} & \textbf{DLinear} & \textbf{TSMixer} & \textbf{MTSMixer} & \textbf{Basisformer} & \textbf{PatchTST} & \textbf{Sundial} & \textbf{TimeMoe} & \textbf{TimesFM} & \textbf{GPT4TS} \\
\hline
1. &\textcolor{blue}{0.602} &0.651 & 1.297 &0.708 &0.824 &0.675 &1.043 &\textcolor{orange}{0.631} &0.665 &0.651 &0.661 &0.677 &1.058 \\
\cline{1-14}
2. &\textcolor{blue}{0.559} &3.824 & 1.187 &0.718 &0.744 &0.639 &0.922 &0.618 &0.615 &0.611 &\textcolor{orange}{0.608} &0.625 &0.987 \\
 \cline{1-14}
3. &\textcolor{blue}{0.572} &4.241 & 1.215 &0.658 &0.763 &0.642 &0.936 &\textcolor{orange}{0.598} &0.638 &0.631 &0.638 &0.640 &1.010 \\
\hline
4. &\textcolor{blue}{0.544} &2.477 & 1.200 &0.684 &0.749 &0.646 &0.958 &\textcolor{orange}{0.573} &0.597 &0.611 &0.628 &0.627 &0.902 \\
\hline
5. &\textcolor{blue}{0.535} &2.732 & 1.163 &0.670 &0.676 &0.610 &0.854 &\textcolor{orange}{0.583} &0.600 &0.599 &0.595 &0.616 &0.914 \\
\hline
6. &\textcolor{blue}{0.494} &2.302 & 1.140 &0.683 &0.636 &0.551 &0.835 &\textcolor{orange}{0.527} &0.552 &0.550 &0.552 &0.564 &0.871 \\
\hline
7. &\textcolor{blue}{0.538} &3.019 & 1.244 &0.753 &0.699 &0.605 &0.920 &0.585 &0.594 &\textcolor{orange}{0.582} &0.621 &0.589 & 0.967\\
\hline
8. &\textcolor{blue}{0.593} &1.901 & 1.154 &0.763 & 0.686&\textcolor{orange}{0.607} &0.822 &0.631 &0.621 &0.631 &0.639 &0.640 &0.942 \\
\hline
9. &\textcolor{blue}{0.577} &1.568 &1.145 &0.743 &0.687 &0.615 &0.811 &\textcolor{orange}{0.611} &0.615 &0.616 &0.616 &0.634 & 0.916\\
 \hline
10. &\textcolor{blue}{0.517} &2.990 & 1.126 &0.808 &0.663 &0.598 &0.790 &\textcolor{orange}{0.559} &0.575 &0.573 &0.580 &0.603 & 0.927\\
\hline
11. &\textcolor{blue}{0.570} &1.739 &1.152 &0.827 &0.682 &0.612 &0.811 &\textcolor{orange}{0.612} &0.619 &0.623 &0.618 & 0.636& 1.001\\
\hline
12. &\textcolor{blue}{0.647} &1.775 & 1.134 &0.831 &0.725 &0.691 &0.787 &0.699 &0.695 &0.719 &\textcolor{orange}{0.674} &0.729 & 1.043\\
\hline
13. &\textcolor{blue}{0.606} &2.156 & 1.294 &0.909 &0.788 &0.643 &1.026 &\textcolor{orange}{0.626} &0.660 &0.676 &0.703 &0.698 & 1.016\\
\hline
14. &\textcolor{blue}{0.641} &1.295 & 1.150 &0.836 &0.795 &0.747 &0.887 &\textcolor{orange}{0.660} &0.703 &0.729 &0.735 &0.754 & 0.950\\
\hline
15. &\textcolor{blue}{0.705} &1.227 & 1.162 &0.859 &0.833 &0.747 &0.956 &\textcolor{orange}{0.726} &0.742 &0.769 &0.775 &0.795 & 0.929\\
\hline

\end{tabular}
}
\caption{Results on 15 out-of-time test subsets for both time series datasets.}
\label{tab:results_wv}
\end{table}
In addition to the results per forecasting sequence length, Table \ref{tab:results_wv} shows that WV-GED \ac{NAS} delivers the lowest RMSE score on average on all 15 test subsets. This highlights the suitability of the MTSMixer-based ensemble for continuous redeployment for energy generation prediction purposes. Furthermore, \ac{NAS} using entropy-based reward produces the worst performance in comparison to all time series models in Table \ref{tab:results_wv}. This is due to the inability of the discovered networks to forecast the setting of 192 hours of total energy generation, which overall inflates the error produced by the entropy-based ensemble on average per out-of-time test subset.\\   

\begin{figure}[h]
\centering
\includegraphics[scale=0.85]{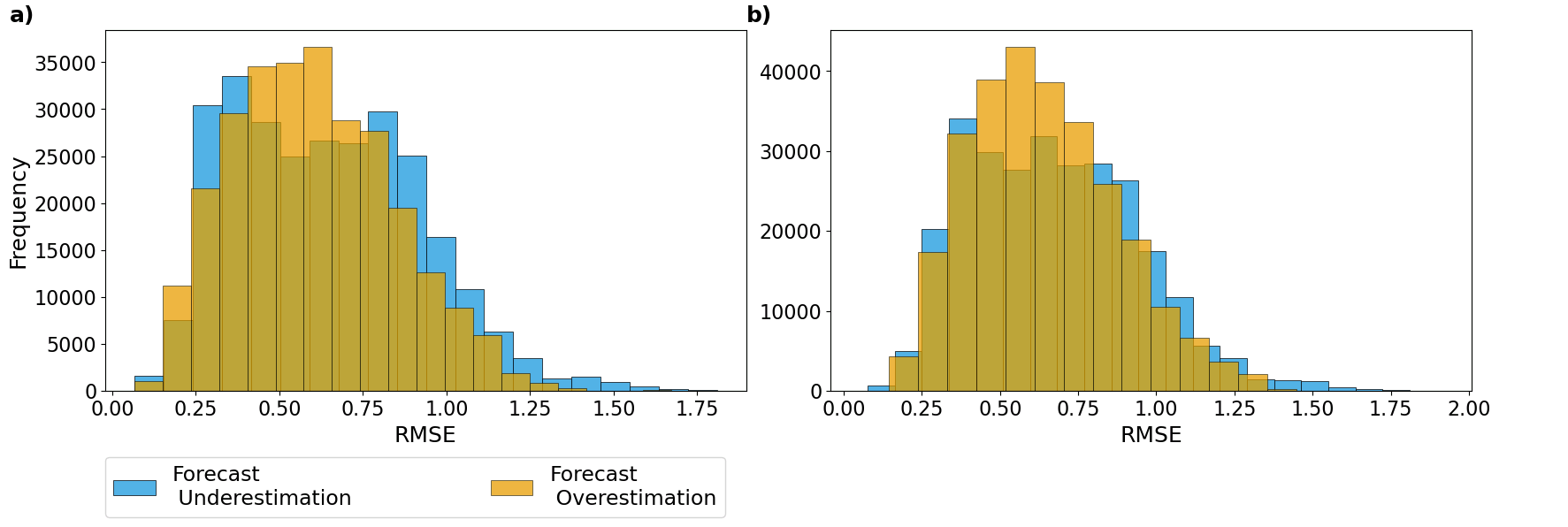}
\caption{Histograms visualizing the RMSE scores associated with underestimated and overestimated forecasts obtained from 10 runs with the models a) \ac{WV}-\ac{GED} \ac{NAS} and b) Basisformer.}
\label{fig:Overestimations_Underestimations}
\end{figure}

\begin{figure}[h]
\centering
\includegraphics[scale=0.75]{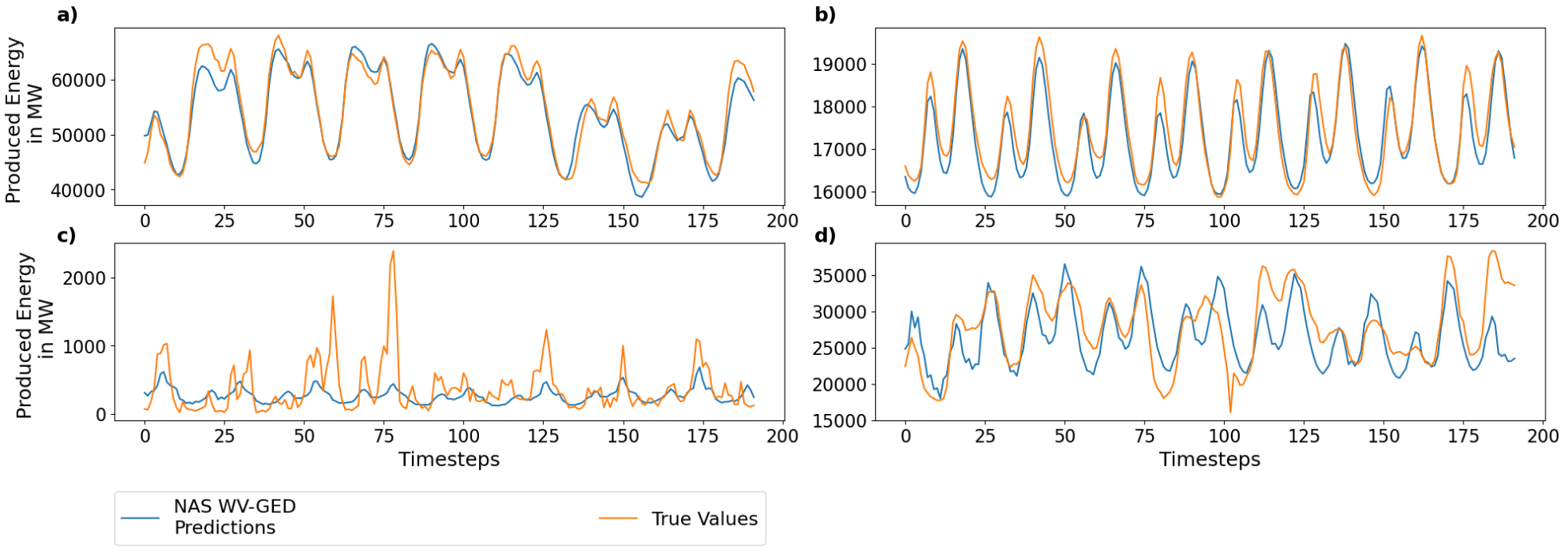}
\caption{True sequence values vs. predictions obtained from \ac{WV}-\ac{GED} \ac{NAS} for the time series associated with a) total supplier generation, b) biomass energy, c) hydro (water reservoir) energy and d) fossil fuel energy.}
\label{fig:Classical_Time_SeriesPlots}
\end{figure}

From an \ac{OR} perspective, the economic implications of \ac{EGF} inaccuracies differ between over- and underestimations of future energy output. Accordingly, Figure \ref{fig:Overestimations_Underestimations} depicts the conditional error distributions associated with both cases for {WV}-\ac{GED} \ac{NAS} and Basisformer, with the latter representing the primary competitor of our approach concerning performance robustness (see Table \ref{tab:results_wv}).
Energy production facilities relying on Basisformer's predictions rather than \ac{WV}-\ac{GED} \ac{NAS}' forecasts are expected to deal with energy deficits more frequently than with energy surplus capacities, as Basisformer’s lower-magnitude errors predominantly stem from overestimations in \ac{EGF}. In addition to certain issues described in Section \ref{subsubsec:SamplingHistory}, e.g., financial losses in energy trading and increased carbon-intensive backup reliance, deploying Basisformer for \ac{EGF} purposes would also amplify the risk of energy under-dispatching. As a result, the necessity for rapid real-time corrections can exacerbate misallocation errors, thereby negatively impacting the overall operational efficiency of energy systems. Concerning \ac{EGF} underestimations, the tails of the right skewed error distribution produced by \ac{WV}-\ac{GED} \ac{NAS} reach lower extreme values than that of Basisformer. Therefore, to elucidate the economic ramifications of \ac{NAS}-based \ac{EGF}, Figure \ref{fig:Classical_Time_SeriesPlots} depicts the deviations between the actual values and 192-hour forecasts across multiple time series.
The higher predictability of the total supplier output in comparison to specific sources underscores the advantage of energy diversification as combining multiple complementary production technologies counteracts the fluctuations of individual generation types. Concerning the risks associated with the latter, the forecasting deviations in Figure \ref{fig:Classical_Time_SeriesPlots} c) highlight \ac{NAS}’ tendency to underestimate the production spikes in renewables such as hydro water reservoir energy. Such fluctuations are likely resulting from varying water availability, which is naturally impacted by changes in weather conditions, e.g., rainfall, solar irradiation, etc. \shortcite{HydroVariability}. While \ac{NAS} is expected to mitigate the mismatch between energy supply and demand more effectively than other approaches, it is unlikely to eradicate the necessity for electricity curtailment operations. The latter deliberately decrease the energy output, e.g., from renewables, to prevent potential overloads, and thus, ensure grid stability \shortcite{EnergyCurtailment_Impact}. Concerning standby sources in the opposite case scenario, i.e., energy deficit, the prediction deviations in Figures \ref{fig:Classical_Time_SeriesPlots} b) and d) showcase the potential for substituting carbon-intensive fossil fuel backup power with environmentally friendly alternatives such as biomass energy.\\

\begin{table}[ht!]
\fontsize{23pt}{23pt}\selectfont
\centering
\renewcommand{\arraystretch}{2.1}
\setlength{\tabcolsep}{4pt}
\resizebox{\textwidth}{!}{
\begin{tabular}{|c|c|*{14}{c|}}
\hline
\multirow{2}{*}{\ip{2.5cm}{\textbf{Time Series \\Model}}} & \multicolumn{4}{c|}{\textbf{Predictive Error}} & \multirow{2}{*}{\ip{3.5cm}{\textbf{Training\\Testing Time}}} & \multicolumn{9}{c|}{\textbf{MCDM Weights for}} &\multirow{2}{*}{\ip{3.0cm}{\textbf{Average\\Model Rank}}}\\
\cline{2-5}
 &  &  & & & & \multicolumn{9}{c|}{\textbf{[Predictive Error, Training Testing Time]}} & \\
\cline{7-15}
 & \textbf{RMSE} & \textbf{MSE} & \textbf{MAE} & \textbf{MedAE} & & \textbf{[0.1, 0.9]}& \textbf{[0.2, 0.8]} & \textbf{[0.3, 0.7]} & \textbf{[0.4, 0.6]} & \textbf{[0.5, 0.5]} & \textbf{[0.6, 0.4]} & \textbf{[0.7, 0.3]} & \textbf{[0.8, 0.2]} & \textbf{[0.9, 0.1]} &  \\
\hline
 \ip{3.0cm}{NAS\\WV-GED} &\textcolor{blue}{0.580}  &\textcolor{blue}{0.459} &\textcolor{blue}{0.471} &\textcolor{blue}{0.411} &0.977 &4 &3 &\textcolor{blue}{2} &\textcolor{blue}{1} &\textcolor{blue}{1} &\textcolor{blue}{1} &\textcolor{blue}{1} &\textcolor{blue}{1} &\textcolor{blue}{1} &\textcolor{blue}{1.667} \\
\hline
TSMixer &0.642  &0.519 &0.528 &0.468 &0.763 &\textcolor{orange}{2} &\textcolor{blue}{1} &\textcolor{orange}{1} &\textcolor{orange}{2} &\textcolor{orange}{2} &\textcolor{orange}{2} &\textcolor{orange}{2} &\textcolor{orange}{2} &3 &\textcolor{orange}{1.889} \\
\hline
DLinear &0.730  &0.670 &0.609 &0.554 &\textcolor{blue}{0.602} &\textcolor{blue}{1} &\textcolor{orange}{2} &3 &3 &4 &4 &5 &6 &7 &3.889 \\
\hline
TimesFM &0.655  &0.594 &0.531 &0.465 &1.691 &5 &5 &4 &4 &3 &3 &4 &5 &3 &4.000 \\
\hline
Basisformer &\textcolor{orange}{0.616}  &\textcolor{orange}{0.500} &\textcolor{orange}{0.503} &\textcolor{orange}{0.444} &2.775 &7 &6 &6 &6 &5 &5 &4 &3 &\textcolor{orange}{2} &4.889 \\
\hline
MTSMixer &0.890  &0.962 &0.760 &0.712 &\textcolor{orange}{0.627} &3 &4 &5 &5 &6 &7 &8 &9 &9 &6.222 \\
\hline
PatchTST & 0.633 &0.526 &0.515 &0.453 &4.210 &9 &8 &8 &8 &7 &6 &6 &5 &4 &6.778 \\
\hline
GPT4TS &0.962 &1.047 &0.814 &0.764 &2.676 &6 &7 &7 &7 &8 &9 &9 &10 &10 &8.111\\
\hline
AutoGluon-TS & 0.763 &0.771 &0.626 &0.559 &5.014 &10 &9 &9 &9 &9 &8 &7 &7 &8 &8.444 \\
\hline
Sundial &0.638  &0.546 &0.519 &0.458 &11.020 &11 &11 &11 &10 &10 &10 &10 &8 &6 &9.667 \\
\hline
\ip{2.5cm}{NAS\\Entropy} &2.263  &77.119 &2.107 &2.033 &1.528 &8 &10 &10 &11 &11 &12 &13 &13 &13 &11.222 \\
\hline
\ip{1.8cm}{Dragon\\NAS} &1.184  &1.678 &1.020 &0.972 &15.100 & 12& 12&12 &12 &12 &11 &11 &11 &12 &11.667 \\
\hline
TimeMoe &0.642  &0.555 &0.525 &0.466 &41.480 &13 &13 &13 &13 &13 &13 &12 &12 &11 &12.556 \\
\hline
\end{tabular}
}
\caption{Overview of Time Series Models Ranking.}
\label{tab:results_ranking}
\end{table}
In the remainder of this section, we provide an overview of the model ranking computed based on four measures quantifying the predictive error, as well as the method runtime measured in minutes (see Table \ref{tab:results_ranking}). All metrics are integrated with different weighting schemes using TOPSIS, i.e., a well-known concept from the field of \ac{MCDM}. TOPSIS quantifies the geometric distance of each metric per model to the best and worst case scenarios in relative terms on a scale from $0 \%\ $ to $100 \%\ $. Overall, our choice for an \ac{MCDM}-based evaluation in the context of time series forecasting is inspired by the application of holistic performance assessment frameworks in other fields, e.g., credit risk modelling \shortcite{C_MCDM}, anomaly detection for network monitoring purposes \shortcite{Anomaly_MCDM}, etc. The ranking presented in Table \ref{tab:results_ranking} for varying \ac{MCDM} weights is estimated based on the resulting TOPSIS score. Since there are four error metrics incorporated into the final model evaluation, an \ac{MCDM} weight of, e.g., 0.1 assigned to the forecasting error is divided equally by four to obtain the weights for each error measure. The remaining 0.9 weight is then assigned to algorithmic efficiency. The weighting schemes in the remaining eight cases are designed analogously. The average model ranking, which incorporates the rankings for the different weighting schemes, highlights the superiority of \ac{NAS} using WV-GED-based reward over the remaining 12 benchmarks. Our approach ranks in the second, third and fourth spots only for three scenarios that assign the highest weights to the model efficiency. Nonetheless, our approach performs very similarly to extremely efficient techniques for global multi-step time series forecasting, such as DLinear, MTSMixer, and TSMixer. This finding underscores the contribution of designing a search space for \ac{NAS} that incorporates only efficient MLP-based computations. The fact that entropy-based \ac{NAS} ranks in the last three spots highlights the significant impact that maximizing the architectural distance and the \ac{WV} penalty can have on the performance of the models discovered with \ac{RL}.

\section{Conclusion and Outlook}
\label{sec:CO}
In this paper, we design a novel \ac{NAS} search space that incorporates efficient components suitable for modeling the intricate temporal patterns of energy-related data. In addition to utilizing SOTA time series blocks borrowed from the field of global multi-step time series forecasting, our \ac{NAS} framework facilitates the automated design of novel periodic-based composite activation functions that enhance the expressive power of the sampled models. Moreover, we formulate a novel reward signal that aims to minimize overfitting in a temporal context and maximize the architectural diversity of the discovered forecasting methods. The multidimensional evaluation of our approach on 15 out-of-time test subsets across three short-term prediction horizon settings unveils the superior performance robustness and the competitive algorithmic efficiency of our \ac{NAS}-based ensemble in comparison to a wide range of time series benchmarks, e.g., from the powerful transformer architectures as well as the recently emerged foundation time series techniques to regression methods designed with previously published \ac{NAS} frameworks. Concerning the latter, the fact that some \ac{NAS} benchmarks rank among the worst performing forecasting techniques highlights the contribution of our \ac{AC}-based search strategy to discover models with high generalization power by sampling solutions from maximally distant regions of the search space.\\

Regarding future directions, our empirical research is limited to \ac{NAS}-based forecasting of energy production time series only. Therefore, it would be useful to explore the performance of our \ac{NAS} framework on other energy-related data, e.g., electricity prices, as well as time series coming from different fields. In addition to chain-structured architectures, future research could also focus on the automated design of forecasting techniques, the components of which are capable of dynamically adjusting to the varying characteristics of temporal data over time, e.g., neural decision trees with different time series blocks in each tree node. Last but not least, an extension of our framework to facilitate the sampling of distributional methods would offer a significant contribution to the modeling of uncertainty in global multi-step time series forecasting.

\newpage
\section*{Acknowledgments}
Stefan Lessmann acknowledges financial support through the project “AI4EFin AI for Energy Finance”, contract number CF162/15.11.2022, financed under Romania’s National Recovery and Resilience Plan, Apel nr. PNRR-III-C9-2022-I8.

\section*{Declaration of generative AI and AI-assisted technologies in the writing process}
 During the preparation of this work, the authors did not use any generative AI tools to produce content for the research article.
 
\bibliographystyle{apacite}
\bibliography{./bib/references.bib} 

\newpage

\appendix
\appendixpage
\begin{appendices}
\section{Actor-Critic Algorithm}
\label{sec:Appendix1}
In this section, we provide a detailed overview of the \ac{AC} model as an example of an \ac{RL} algorithm.\\

A \ac{MDP} characterizes a system that models the interactions between an agent and an environment. The latter produces rewards associated with new states as a result of the controller network, i.e., the agent, taking certain actions. These actions aim to maximize the rewards collected by the agent in the long term. Therefore, \ac{MDP}s play a central role in \ac{RL}-based applications, where the controller network learns an optimal policy based on the feedback from the environment. The \ac{AC} model represents an on-policy \ac{RL} algorithm, which incorporates a policy-based and a value-based component. The actor network is associated with the action-selection policy, which facilitates the mapping of states to sampled actions. The critic network produces the state-value function, which estimates the cumulative future reward that can be obtained from the current state of the environment. The advantage function, which highlights the connection between the actor and the critic networks, is expressed in the following way:\\
\begin{equation}
\qquad\qquad\qquad\qquad\qquad\qquad\qquad
A^\pi_{t}=Q^\pi_{t}(s_{t\minus1},a_{t})\minus V^\pi_{t}(s_{t\minus1})
\label{eq:advantage_function}
\end{equation}
where $\pi$ is the policy of the agent in the current time step $t$, $Q^\pi_{t}(s_{t\minus1},a_{t})$ is the action-value function, which measures the quality of the selected action $a_{t}$ based on the environment state $s_{t\minus1}$ produced in the previous time step, and $V^\pi_{t}(s_{t\minus1})$ is the state-value function produced by the critic. Overall, positive values of $A^\pi_{t}$ indicate that the action selected by the policy network in the current time step is better than any other action that could have been selected. By contrast, negative values of $A^\pi_{t}$ imply that the critic's expectation for taking a certain action was higher than its true estimated value. Since the advantage function acts as a scaling factor for the gradients of the policy network, positive (negative) values of $A^\pi_{t}$ increase (decrease) the probabilities for taking the same action in the future. The $Q$-function quantifies each sampled action's quality by computing the discounted sum of rewards to go:\\
\begin{equation}
\qquad\qquad\qquad\qquad\qquad\qquad\qquad
Q^\pi_{t}=\big( \sum_{t^\prime=t}^{T} \gamma^{t^\prime \minus t} R_{t^\prime}\big)
\label{eq:discounted_sum_rewards}
\end{equation}
where $\gamma$ is the discount factor, which we set to 0.99, and $t^\prime \minus t$ is the exponent, which reduces the weight of the rewards achieved in each following time step $t^\prime$. Therefore, the $Q$-function accounts for rewards that are obtained only in and after the current time step of an episode with length $T$. This is based on the assumption that the current time point can have an impact on the future, whereas future time steps cannot influence the past. Additionally, events that occur far in the future are associated with higher uncertainty than currently occurring events or events that occur in the near future. The high uncertainty is associated with a high probability that future rewards would have high variance. Therefore, the further we go into the future within a single episode, the more we discount the obtained rewards.  Both the subtraction of $V^\pi_{t}(s_{t\minus1})$ from the validation rewards, as shown in Equation \ref{eq:advantage_function}, and the discounting of the rewards, as shown in Equation \ref{eq:discounted_sum_rewards}, contribute to reducing the variance of the actor's gradients.  This prevents the agent from taking large steps in the wrong direction. During the backward pass, the actor network's gradients are weighted with the advantage function. Overall, the actor aims to maximize the probability of the most promising actions sampled over time, whereas the critic aims to minimize the advantage function. Thus, the critic improves in guiding the actor's decisions during the sampling process. 


\newpage
\section{\ac{NAS} Sampling History}
\label{sec:Appendix2}
This section presents details about the differences in the sampling history of the \ac{AC} networks trained with our novel reward signal and the entropy-based objective.\\

\begin{figure}[h]
\centering
\includegraphics[scale=0.95]{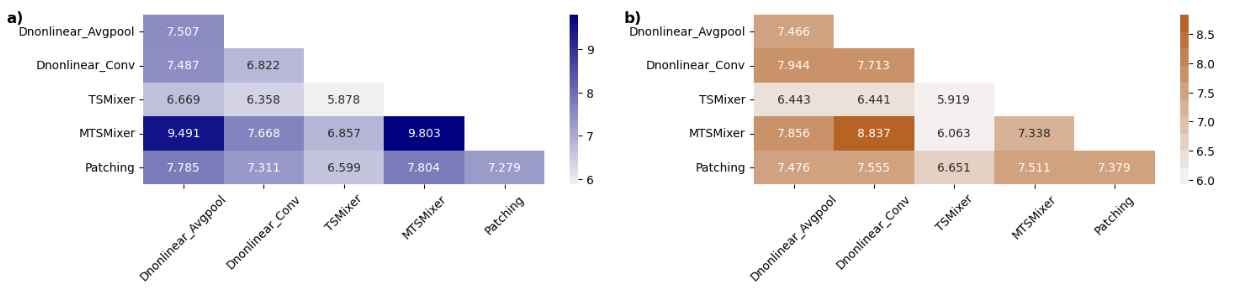}
\includegraphics[scale=0.95]{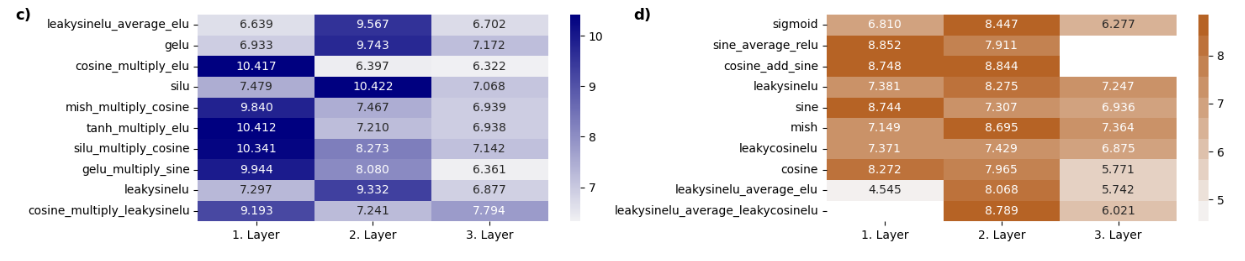}
\includegraphics[scale=0.7]{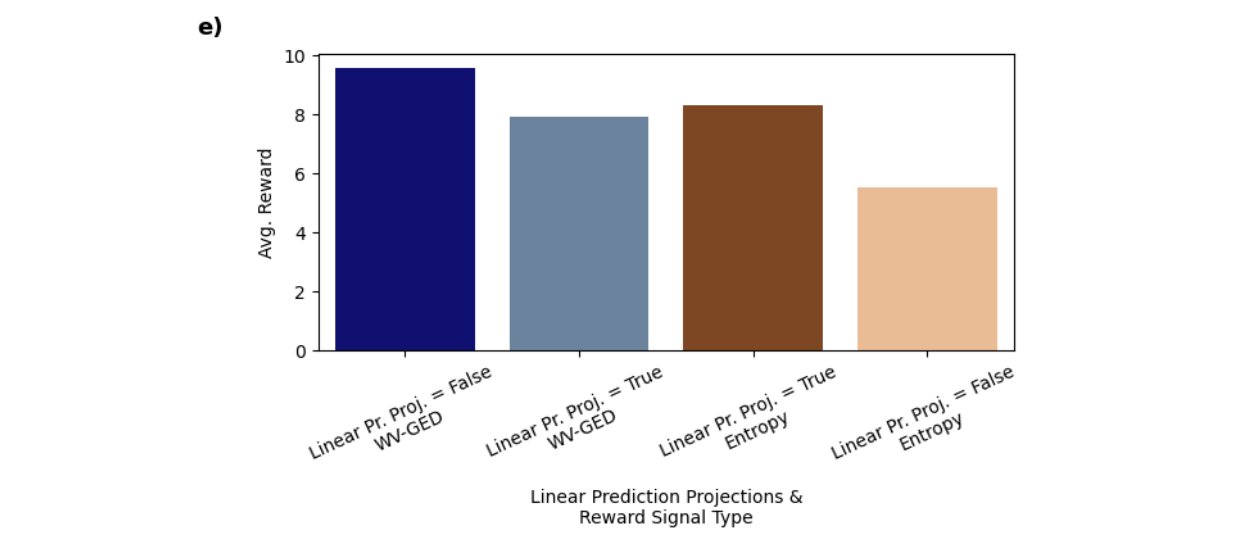}
\caption{Overview of the average reward associated with certain architecture components of the \ac{NAS} models sampled with the two controller networks, a) and b) pair combinations of time series blocks sampled during \ac{WV}-\ac{GED} \ac{NAS} and entropy \ac{NAS}, c) and d) (composite) nonlinearities per hidden layer, e) application of linear prediction projections following the hidden time series layers.}
\label{fig:NAS_SamplingHistory_Details}
\end{figure}





\end{appendices}

\end{document}